\documentclass[letterpaper]{article} 
\usepackage{aaai2027}
\usepackage[hyphens]{url}  
\usepackage{graphicx} 
\usepackage{natbib}  
\usepackage{caption} 
\usepackage{amsmath}
\usepackage{amssymb}
\usepackage{booktabs}
\usepackage{algorithm}
\usepackage{algorithmic}
\newtheorem{proposition}{Proposition}
\newtheorem{lemma}{Lemma}

\nocopyright

\title{Verify, Repair, Repeat, or Stop? Robust Stopping for Noisy Verify-Repair Loops in LLM Agents}
\author{
    Yitao Wu\textsuperscript{\rm 1},
    Si Shen\textsuperscript{\rm 2},
    Rui Yang\textsuperscript{\rm 3}\thanks{Corresponding author.},
    Hong Peng\textsuperscript{\rm 3},
    Bin Hu\textsuperscript{\rm 3}
}
\affiliations{
    \textsuperscript{\rm 1}saofund.ai, Shenzhen, China\\
    \textsuperscript{\rm 2}Shenzhen Xingqing Zhiti Technology Co., Ltd., Shenzhen, China\\
    \textsuperscript{\rm 3}School of Information Technology and Engineering, Lanzhou University, Lanzhou, China\\
    yitaowu123@gmail.com, sishen4@acm.org, rykeryang@163.com, pengh@lzu.edu.cn, bh@lzu.edu.cn
}

\begin{document}

\maketitle

\begin{abstract}
Verify-repair loops are a standard means for large language model (LLM) agents to correct faulty plans in code generation, mathematical reasoning, and tool use. When both the verifier and the repairer are noisy, repair can damage already-correct plans, and reported acceptance keeps rising while true validity falls, so existing methods lack a principled basis for deciding when repair should stop. We propose VRR-Stop, a robust stopping framework for noisy verify-repair-repeat (VRR) loops. A four-parameter noise model separates verifier false acceptance and false rejection from the repair and damage behavior of the repairer. Belief filtering turns repeated verification votes into an estimate of committed validity, and the loop commits or repairs according to the sign of the true marginal gain, which requires only sign identifiability rather than accurate recovery of all parameters. When verifier discrimination approaches zero, calibration itself fails and estimation error can flip the stopping sign, so we pair VRR-Stop with VRR-Guard, an estimation-free fallback that replaces the incumbent candidate only under a sufficient verification margin. On a GSM8K stress setting, VRR-Stop improves final true validity by 60.6 percentage points over fixed five-round repair at an average cost of 0.72 repair rounds. Across settings, stopping reliability is governed jointly by verifier discrimination and the decision margin rather than by the absolute size of estimation error.
\end{abstract}

\begin{links}
    \link{Code}{anonymous.4open.science/r/vrr-artifact-2583}
\end{links}

\section{Introduction}
\label{sec:intro}

On complex multi-step tasks such as code generation, tool invocation, and web automation, a single plan generated by an LLM agent often contains logical gaps, constraint violations, or failing tool calls, and once such errors enter execution they can derail the entire task. Verify-repair loops are therefore widely adopted, in which a verifier checks the candidate plan, a repairer produces a revision from the feedback, and the loop iterates toward a valid solution. A large body of work shows that this feedback mechanism corrects explicit errors and significantly improves plan executability and reasoning quality \citep{madaan2023selfrefine,shinn2023reflexion,lightman2023lets}. In practice, however, the paradigm implicitly relies on two premises that do not always hold, namely that multi-round repair tends to improve true quality and that verifier outputs suffice to reflect true validity. When both the verifier and the repairer are noisy, the two premises fail simultaneously, turning the gains of the loop into an unpredictable systemic risk.

\begin{figure}[t]
  \centering
  \includegraphics[width=0.85\columnwidth]{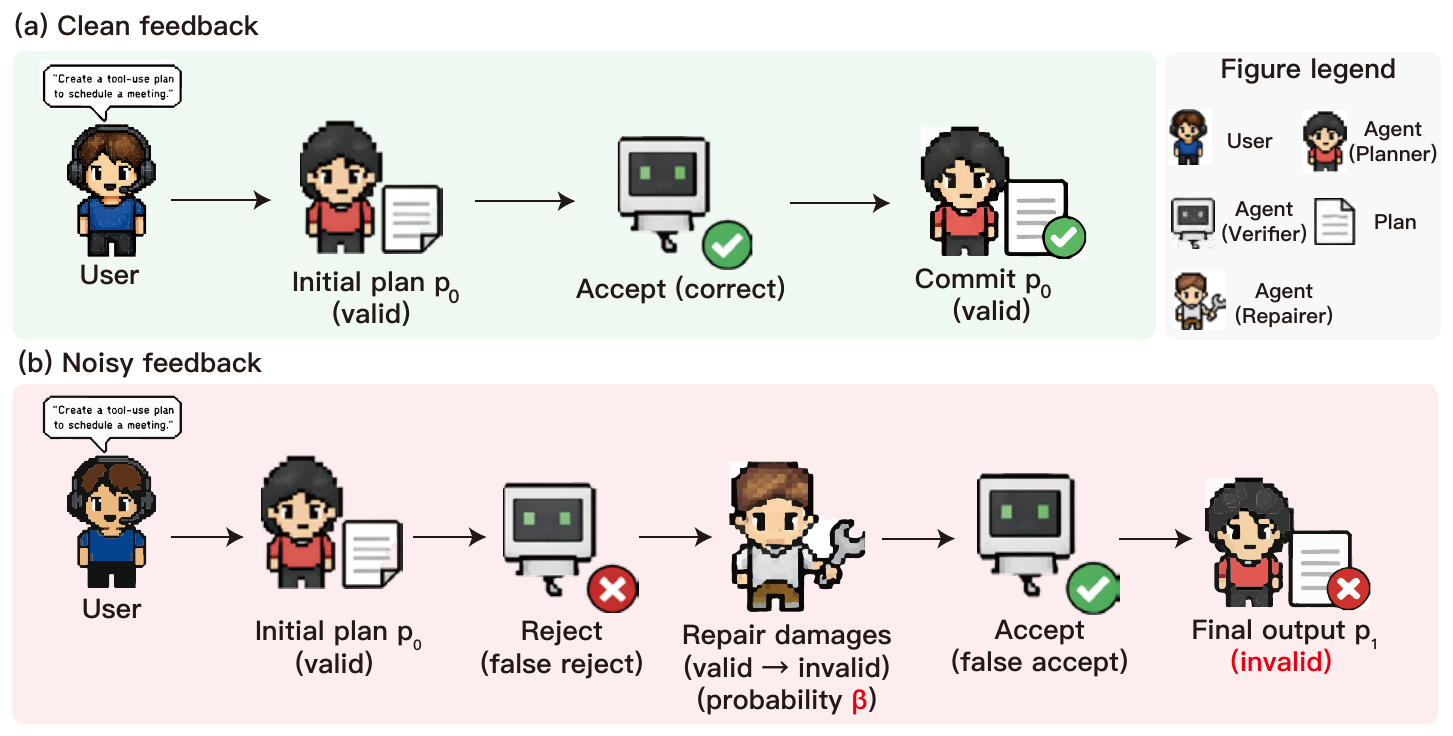}
 \caption{
Clean feedback accepts and commits the valid plan \(p_0\), whereas noisy feedback can damage it into an invalid \(p_1\).
Verifier and repair errors can therefore compound, making continued repair harmful.
}
  \label{fig:motivation}
\end{figure}

Figure~\ref{fig:motivation} illustrates a typical instance of this mismatch. A truly valid initial plan \(p_0\) can be falsely rejected by the verifier under noisy feedback, turned into an invalid plan \(p_1\) by a harmful repair, and finally falsely accepted by the verifier. Verifier errors and repairer errors thus compound inside the loop, so true quality can decline over rounds instead of improving. Prior studies likewise observe that proxy signals in self-repair, LLM judges, and inference-time search can decouple from true quality \citep{huang2023large,kamoi2024evaluating,pan2024feedback,pan2024spontaneous,khalaf2025inference}. Verification pass rates therefore cannot be equated with the true validity of plans, and multi-round repair does not guarantee monotone improvement. The central question of this paper is therefore how, in agent systems that rely on verify-repair loops, one can decide when repair has degenerated from improving true quality into fitting verifier noise.

Existing feedback-driven research can be examined from three directions with respect to this stopping problem. Self-correction and reflective feedback let the model revise its output from its own feedback, verbal reflection, or the environment \citep{madaan2023selfrefine,shinn2023reflexion,kamoi2024when}, yet verifier feedback and repairer behavior remain entangled in the loop, and pass-rate fluctuations cannot be attributed to either source. How to separate verifier noise from repairer noise, and to characterize how true validity evolves with repair rounds, is a key \emph{loop-noise modeling} problem. Process supervision, external verifiers, and generative reward models constrain intermediate reasoning with explicit checking signals \citep{cobbe2021training,lightman2023lets,zhang2024generative,khalifa2025process}, but their effectiveness presupposes reliable feedback. When the feedback itself is noisy, how to estimate parameters from weak supervision while guaranteeing that estimation error does not turn ``should stop'' into ``should continue'' is a difficult \emph{stopping identifiability} problem. Studies of test-time search and proxy-signal optimization show that imperfect verifiers, false acceptance, self-bias, and reward over-optimization can limit or even reverse inference-time gains \citep{gao2023scaling,stroebl2024limits,yu2025scaling,dorner2025rocnreroll,zhou2025evaluating,lu2025when}. How to detect such failure in time, and to design a conservative stopping mechanism that does not rely on accurate parameter estimation so that continued repair no longer causes harm, is an urgent \emph{safety fallback} problem.

Targeting the basic dilemma that continued repair does not always improve true quality, we start from the three problems of loop-noise modeling, stopping identifiability, and safety fallback, and propose \textbf{VRR-Stop}, a robust stopping framework for noisy verify-repair-repeat (VRR) loops. The contributions of this paper are as follows.
\textbf{(i)} We propose a four-parameter loop-dynamics model that explicitly separates verifier false acceptance, verifier false rejection, and the repair and damage behavior of the repairer, thereby characterizing how the true validity of plans evolves over multiple repair rounds.
\textbf{(ii)} We derive a stopping criterion driven by the sign of the true marginal gain, and give a weakly supervised estimation method based on multi-round verification records and a small number of labeled transition samples. The stopping decision requires only that the gain sign be identifiable rather than accurate recovery of all parameters.
\textbf{(iii)} We give an explicit criterion for stopping-sign identifiability, show that decision reliability is jointly determined by the verifier discrimination $J$ and the decision margin $\Delta$, and empirically expose this boundary in low-discrimination verifier scenarios.
\textbf{(iv)} We design VRR-Guard, a conservative fallback that does not rely on accurate parameter estimation and avoids fixed-round-repair-style degradation under sign unidentifiability or distribution shift.

\section{Related Work}
\label{sec:related}

\textbf{Iterative self-correction.}
Self-Refine revises outputs with self-generated feedback, Reflexion stores verbal feedback as experience, CRITIC obtains verifiable critiques from external tools, and SCoRe trains self-correction with reinforcement learning \citep{madaan2023selfrefine,shinn2023reflexion,gou2024critic,kumar2024training}. Follow-up studies find that, without reliable external signals, models may fail to recognize their own errors and can corrupt previously correct answers \citep{huang2023large,kamoi2024evaluating,kamoi2024when}. This line focuses on producing better feedback and revisions, but does not formalize termination as a decision problem in which the verifier and the repairer are simultaneously noisy. VRR-Stop separates verifier false acceptance and false rejection from repair and damage behavior, and stops according to the true marginal gain of continued repair.

\textbf{Verifier-guided inference and proxy over-optimization.}
Verifier-guided methods score candidates with outcome verifiers, process reward models, or generative verifiers for candidate selection, search, or further reasoning \citep{cobbe2021training,lightman2023lets,zhang2024generative,khalifa2025process}, with recent work scaling sampling and verification budgets \citep{zhao2025sample,zhong2025solvedetectverify}. The gains are bounded by verifier coverage, ranking error, and ROC characteristics \citep{stroebl2024limits,yu2025scaling,dorner2025rocnreroll,lu2025when}, and proxy scores in feedback loops decouple from true quality through training- and inference-time reward hacking \citep{pan2024feedback,pan2024spontaneous,gao2023scaling,khalaf2025inference}. HedgeTune \citep{khalaf2025inference} tunes the operating point of selection-style mechanisms, and ROC-n-reroll \citep{dorner2025rocnreroll} analyzes how imperfect verifiers limit resampling. Unlike these selection-style mechanisms over independent candidates, a verify-repair loop keeps rewriting the same candidate, so the repairer can both fix invalid plans and damage valid ones. VRR-Stop targets the state transitions and stopping decision of this path-dependent rewriting rather than re-establishing proxy over-optimization itself.

\textbf{Self-correction dynamics and reliable stopping.}
A probabilistic theory of self-correction describes accuracy converging to a fixed ceiling at a single rate \citep{yang2025probabilistic}, which cannot express the interior peaks and late-stage decline caused by joint verifier and repairer noise. Unlabeled accuracy estimation infers judge reliability from multiple classifiers \citep{platanios2016estimating}, Youden's $J$ measures how well a judge preserves class differences \citep{collot2025balanced}, and uncertainty-aware process verification flags unreliable reward-model steps \citep{ye2025uncertainty}. More recent work treats stopping itself as a test-time decision dimension, via adaptive early stopping and confidence-based trace filtering for chain-of-thought reasoning \citep{sun2026stop,fu2026deepconf} or statistically valid evaluation under imperfect judges \citep{feng2026noisy}, but these methods address single-pass generation or independent sampling and do not model the true-state transitions induced by repair. VRR-Stop uses Youden's $J$ for stopping identifiability rather than static judge ranking, and switches to the guarded keep-best fallback when the sign is unidentifiable.

\section{Preliminaries and Motivation}
\label{sec:prelim}

\subsection{Verify-Repair Loops}
\label{ssec:loops}

We consider a canonical verify-repair loop. Given a task instance, the agent first generates an initial plan $p_0$. In round $k$, the system issues $M$ independent verification queries on the current plan $p_k$; each query returns a binary signal $a_k^{(m)}\in\{0,1\}$ ($1$ for accept), yielding the acceptance count $S_k=\sum_{m=1}^{M}a_k^{(m)}\in\{0,\ldots,M\}$. Based on $S_k$ and the history, the system updates the committed validity $b_k=\Pr(y_k=1\mid H_k)$ and decides whether to commit the current plan or to repair it. The observables and the decision variable of a single round are given in Eq.~\eqref{eq:verify-repair-loop}:
\begin{equation}
\label{eq:verify-repair-loop}
S_k = \sum\nolimits_{m=1}^{M} a_k^{(m)},\qquad
\pi(H_k) \in \{\textsc{Commit}, \textsc{Repair}\},
\end{equation}
where $\pi(H_k)$ is the stopping policy over the observable history $H_k$. If $\pi(H_k)=\textsc{Commit}$, the loop terminates and commits $p_k$; otherwise the repairer produces the next plan $p_{k+1}$ from the current plan and the verification feedback.

To separate observable verifier outputs from unobservable true quality, let $y_k\in\{0,1\}$ denote the true validity of $p_k$, i.e., whether the plan actually satisfies the task constraints, is executable, and completes the goal. The observable history up to round $k$ is defined in Eq.~\eqref{eq:observed-history}:
\begin{equation}
\label{eq:observed-history}
H_k=(p_0,S_0,\ldots,p_k,S_k),\qquad y_k\in\{0,1\}.
\end{equation}
Since $y_k$ is generally unavailable at deployment, the stopping policy must act on $H_k$ alone. An accepted plan is therefore not necessarily valid, and a rejected plan is not necessarily invalid.

\subsection{Noise Model}
\label{ssec:noise}

Observed signals do not equal true validity, and repair is not a monotone improvement process. To capture both kinds of uncertainty, we adopt a compact noise model, given in Eq.~\eqref{eq:noise-model}:
\begin{equation}
\label{eq:noise-model}
\begin{aligned}
\rho_0 &= \Pr(a_k^{(m)}=1 \mid y_k=0),\\
\rho_1 &= \Pr(a_k^{(m)}=0 \mid y_k=1),\\
\alpha &= \Pr(y_{k+1}=1 \mid y_k=0,\mathrm{repair}),\\
\beta  &= \Pr(y_{k+1}=0 \mid y_k=1,\mathrm{repair}),\\
J &= 1-\rho_0-\rho_1 ,
\end{aligned}
\end{equation}
where $\rho_0$ is the probability that an invalid plan is falsely accepted, $\rho_1$ the probability that a valid plan is falsely rejected, $\alpha$ the probability that repair turns an invalid plan valid, and $\beta$ the probability that repair damages a valid plan. $J$ is the discrimination ability of the verifier. As $J$ approaches $0$, acceptance signals can no longer support fine-grained stopping decisions.

The model treats repeated queries on the same plan as conditionally independent observations given the true state, and treats $\rho_0,\rho_1,\alpha,\beta$ as stable within a local decision window. The former supports Bayesian updates from acceptance counts; the latter lets the one-step marginal gain be characterized by compact parameters. The conditional-noise assumption also ignores instance-level difficulty heterogeneity. The same verifier may err at different rates on different plans, and repeated queries may be correlated; a beta-binomial estimator diagnoses such within-class heterogeneity (Appendix~E). Both assumptions are local approximations rather than global stationarity claims; their applicable range under non-stationarity is analyzed in the experiments.

\subsection{Stopping Objective}
\label{ssec:objective}

The question is when, given $H_k$, the system should repair once more rather than commit. We define the true marginal gain of one more repair round as the expected change in true validity, and stop when it no longer exceeds a minimum-gain threshold, as specified in Eq.~\eqref{eq:stopping-objective}:
\begin{equation}
\label{eq:stopping-objective}
\begin{aligned}
G_k &= \mathbb{E}\!\left[y_{k+1}-y_k \mid H_k,\mathrm{continue}\right],\\
\pi(H_k) &=
\begin{cases}
\textsc{Repair}, & G_k>\tau,\\
\textsc{Commit}, & G_k\le \tau ,
\end{cases}
\end{aligned}
\end{equation}
where $\tau\ge 0$ converts extra computation into a minimum acceptable validity gain or acts as a conservative margin; $\tau=0$ when cost is ignored. Eq.~\eqref{eq:stopping-objective} adopts the one-step gain as the stopping target, comparing only a commit now against exactly one more repair round, without look-ahead over multi-round trajectories. Since $y_k$ is unobservable at deployment, Eq.~\eqref{eq:stopping-objective} only defines the target; estimating $G_k$ is deferred to the method section.

\subsection{Motivation}
\label{ssec:motivation}

\begin{figure}[t]
    \centering
    \includegraphics[width=0.80\linewidth]{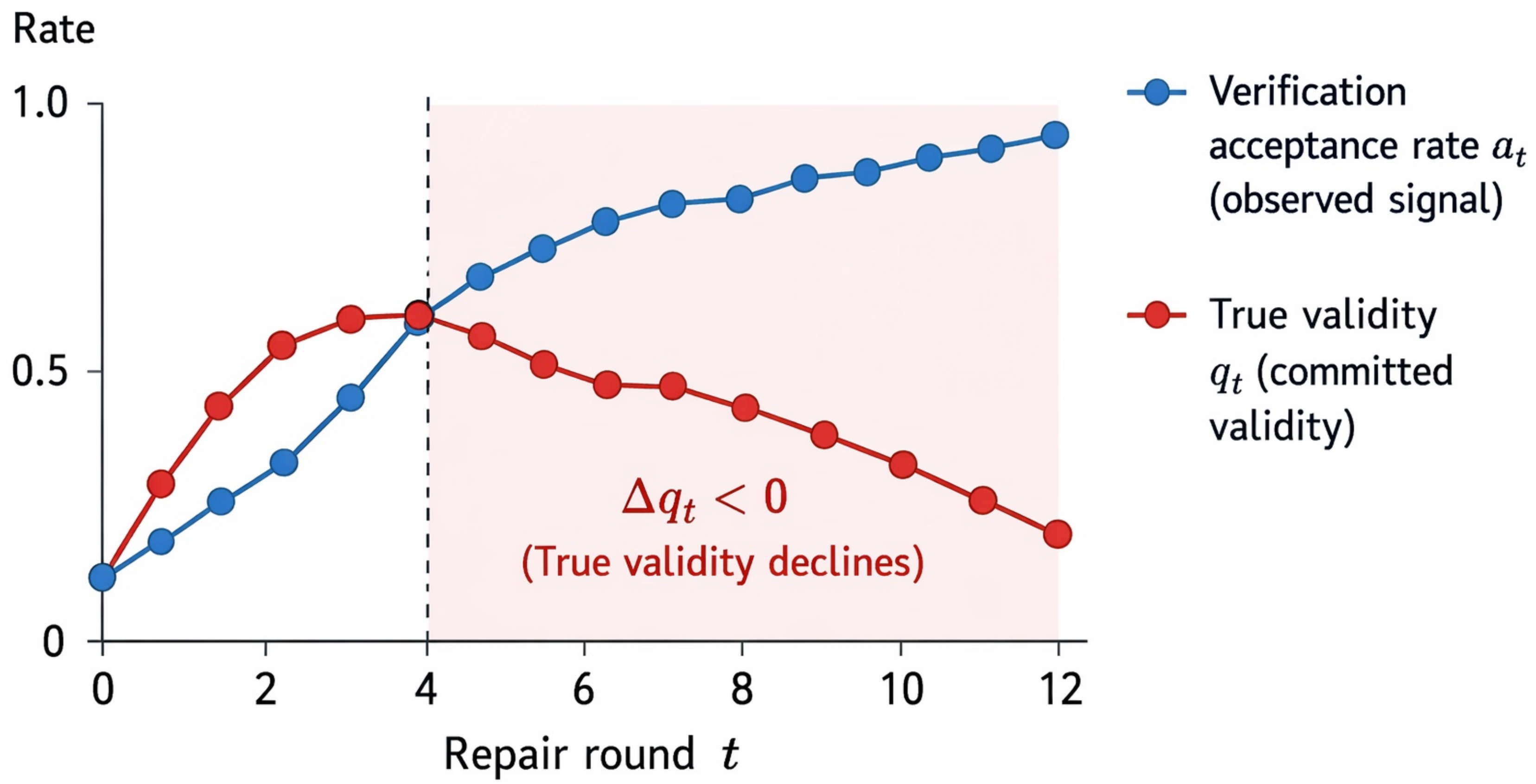}
   \caption{
The population verification rate \(\bar A_t\) can keep increasing while the true-validity rate \(Q_t\) peaks and then declines.
The shaded interval marks \(\Delta Q_t<0\), where continued repair becomes harmful.
}
    \label{fig:signal-quality-decoupling}
\end{figure}

\textbf{More repair is not always beneficial.}
When the damage probability is non-negligible, simply adding repair rounds can push true quality past a peak into sustained decline. Iterative refinement methods such as Self-Refine, Reflexion, and CRITIC assume that outputs improve over rounds of feedback \citep{madaan2023selfrefine,shinn2023reflexion,gou2024critic}. Let $Q_t=\Pr(y_t=1)$ be the population true-validity rate (subscript $t$ indexes population-level round dynamics, distinct from the per-instance index $k$). If every round triggers repair, then $Q_{t+1}=Q_t(1-\beta)+(1-Q_t)\alpha$, so whenever $Q_t$ is already high or $\beta$ is non-negligible, $Q_{t+1}-Q_t\le 0$ and repair noise accumulates over rounds. As Fig.~\ref{fig:signal-quality-decoupling} shows, $Q_t$ enters a declining regime after $t^\star$, the round at which the marginal gain turns negative. This risk is not a corner case under extreme parameters. Self-correction is known to corrupt correct answers \citep{huang2023large,kamoi2024evaluating}, and our measured repair trajectories reach $\beta$ close to $0.94$. Multi-round repair therefore cannot be treated as a safe default; an explicit stopping mechanism is required.

\textbf{Verifier signals can improve without true validity improving.}
Treating verifier pass rates as a quality metric makes the system overestimate the benefit of repair \citep{zhao2025sample,zhong2025solvedetectverify,khalifa2025process}. With $\bar A_t=\mathbb{E}[S_t/M]$, the population verification rate satisfies $\bar A_t=\rho_0+(1-\rho_0-\rho_1)Q_t$, where $1-\rho_1$ and $\rho_0$ are the verifier true- and false-positive rates. When $J=1-\rho_0-\rho_1$ is low or the noise parameters drift, the system can exhibit $\bar A_{t+1}-\bar A_t>0$ while $Q_{t+1}-Q_t\le 0$: the dashed curve in Fig.~\ref{fig:signal-quality-decoupling} keeps rising while true validity already falls. Verifier calibration can degrade quickly after switching models, verifiers, or task distributions, so optimizing pass rates alone does not guarantee quality gains; verifier noise and true validity must be modeled separately.

\textbf{Stopping requires identifiable evidence.}
Heuristic stopping rules such as fixed round budgets, consecutive-pass counts, or score thresholds cannot reliably judge whether repair is still worthwhile. Existing test-time scaling and verifier-guided methods focus on exploiting more sampling, verification, or repair, and rarely model whether the parameters required for stopping are identifiable under weak supervision. What the decision actually needs is the sign of $\Delta Q_t=Q_{t+1}-Q_t$, while the available signals are only the acceptance counts $\{S_0,\ldots,S_t\}$ and a few labeled repair transitions. If estimation error flips the sign of $\widehat{\Delta Q_t}$, the system mistakes ``should stop'' for ``should continue'' and enters a regime of sustained damage. Class balance, verifier calibration, and Youden's $J$ all affect judge reliability, and accuracy estimation without labels itself requires assumptions \citep{collot2025balanced,platanios2016estimating}. A stopping policy should therefore be near-optimal when evidence suffices and fall back to a conservative decision when it does not.

\section{Method}
\label{sec:method}

\begin{figure*}[t]
  \centering
    \includegraphics[width=0.93\textwidth]{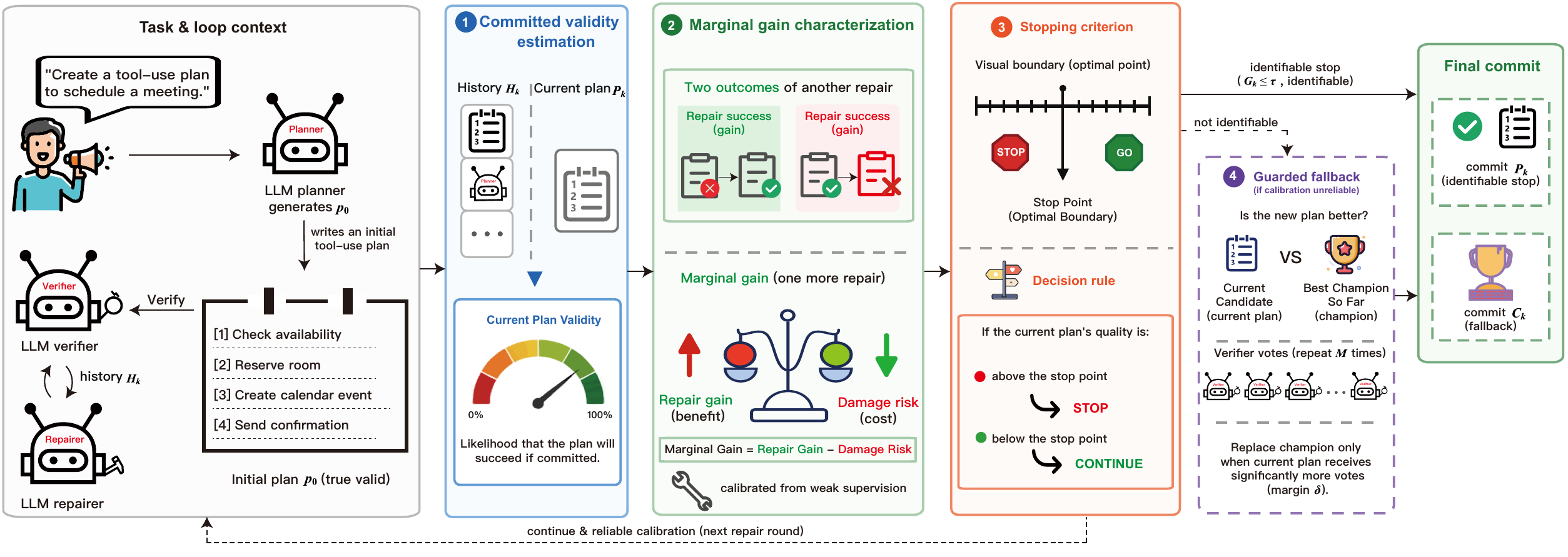}
  \caption{Overview of VRR-Stop, which estimates committed validity \(b_k\) and stops according to the marginal gain \(G_k\). VRR-Guard is evaluated separately as a conservative fallback when calibration is unreliable.}
  \label{fig:method-overview}
\end{figure*}

Figure~\ref{fig:method-overview} gives an overview. Given the verify-repair history $H_k$ and one-off weakly supervised calibration data, the framework estimates the committed validity of the current plan round by round and outputs commit or repair. It consists of a belief estimator that turns the $M$ verification votes of each round into the posterior $b_k$ (stage one), a stopping criterion that forms the marginal gain $G_k$ and acts on its sign (stages two and three), and the conservative fallback VRR-Guard that takes over commitment when calibration is unreliable (stage four). The three components address, in order, the loop-noise modeling, stopping identifiability, and safety fallback problems raised in the introduction.

\subsection{Belief Filtering and the Stopping Boundary}
\label{ssec:belief}

Given $H_k$, the committed validity $b_k=\Pr(y_k=1\mid H_k)$ is the posterior probability that the current plan is truly worth committing; it is not the verification pass rate, which can rise merely through false acceptance.

The belief evolves by an update--predict recursion. The initial belief is the generator prior $b_0^{-}=\hat\pi$, estimated on the calibration folds as the true-validity rate of initial plans. In round $k$, the $M$ queries are conditionally independent given $y_k$, so $S_k$ is binomial and the observation update is given in Eq.~\eqref{eq:belief-update}:
\begin{equation}
\label{eq:belief-update}
b_k=\frac{b_k^{-}(1-\rho_1)^{S_k}\rho_1^{M-S_k}}
{b_k^{-}(1-\rho_1)^{S_k}\rho_1^{M-S_k}+(1-b_k^{-})\rho_0^{S_k}(1-\rho_0)^{M-S_k}},
\end{equation}
where $b_k^{-}$ is the predictive belief before observing this round's votes; the binomial coefficients cancel. If repair is executed, the true state transitions by $\alpha,\beta$ and the predict step is $b_{k+1}^{-}=(1-\beta)b_k+\alpha(1-b_k)$. The recursion relies on one explicit assumption, namely that the repair transition is conditionally independent of $H_k$ given $y_k$. Although plans entering repair are filtered by verification signals, their transitions are still governed by $\alpha,\beta$; the labeled transition samples are collected from frozen trajectories in which every round triggers repair, consistent with this assumption.

Subtracting the current belief from the predictive belief gives the true marginal gain of one more repair round, and setting it to zero yields the critical posterior belief, as in Eqs.~\eqref{eq:gated-gain}--\eqref{eq:gated-boundary}:
\begin{align}
G_k&=b_{k+1}^{-}-b_k=(1-b_k)\alpha-b_k\beta,\label{eq:gated-gain}\\
b^\star&=\alpha/(\alpha+\beta),\qquad \alpha+\beta>0,\label{eq:gated-boundary}
\end{align}
where the first term of Eq.~\eqref{eq:gated-gain} is the expected gain from fixing an invalid plan and the second the expected loss from damaging a valid one. $b^\star$ is exactly the fixed point of the population dynamics $Q_{t+1}=Q_t(1-\beta)+(1-Q_t)\alpha$ of the motivation subsection. Repair can only push validity toward $b^\star$, and once the belief crosses it, the expected direction of further repair is downward. The belief space thus splits into a repair-beneficial regime ($b_k<b^\star$) and a repair-harmful regime ($b_k\ge b^\star$). The boundary depends only on the repairer's $\alpha$ and $\beta$; verifier noise $\rho_0,\rho_1$ does not move the boundary but determines, through $b_k$, whether the comparison is reliable (next subsection).

Stopping follows the one-step objective of Eq.~\eqref{eq:stopping-objective} with $G_k$ from Eq.~\eqref{eq:gated-gain}. All experiments use $\tau=0$; a nonzero threshold shifts the boundary to $b_\tau^\star=(\alpha-\tau)/(\alpha+\beta)$, and no repair is issued if $\alpha\le\tau$. The rule ties the stopping time to a direct comparison between repair benefit and damage risk rather than to a preset budget or pass-rate threshold. When the round budget $K_{\max}$ is exhausted, the current plan is committed.

\subsection{Calibration and Sign Identifiability}
\label{ssec:method-calibration}

At deployment, none of $b_k,\rho_0,\rho_1,\alpha,\beta$ is directly observable. Calibration uses two kinds of weak supervision. For repeated verification records, per-plan acceptance counts are modeled as a two-component binomial mixture whose components correspond to truly valid and invalid plans; a binomial-mixture expectation-maximization (EM) estimator recovers the mixture weight and $\hat\rho_0,\hat\rho_1$ without labels. For repair transitions, $\hat\alpha,\hat\beta$ are frequency estimates from at most 300 labeled before-after plan pairs, with labels produced by each task's validity-judging rule.

Substituting the estimates into Eq.~\eqref{eq:gated-gain} gives $\hat G_k=(1-\hat b_k)\hat\alpha-\hat b_k\hat\beta$, with $\hat b_k$ computed from Eq.~\eqref{eq:belief-update} under $\hat\rho_0,\hat\rho_1$, so calibration error propagates into $\hat G_k$. Let the error radius $B_k$ satisfy $\Pr(|\hat G_k-G_k|\le B_k)\ge 1-\eta$, where $\eta$ is a user-chosen tolerance; $B_k$ shrinks as the calibration sample size, the verification budget $M$, and the discrimination $J$ grow. The sign-identifiability condition is given in Eq.~\eqref{eq:sign-identifiability}:
\begin{equation}
\label{eq:sign-identifiability}
\hat{G}_k-B_k>\tau
\quad \text{or} \quad
\hat{G}_k+B_k\le \tau.
\end{equation}

\begin{proposition}[Stopping-sign consistency]
\label{prop:sign-consistency}
If Eq.~\eqref{eq:sign-identifiability} holds, the action taken under $\hat G_k$ agrees with the action under the true $G_k$ with probability at least $1-\eta$.
\end{proposition}

The proof and a sample-complexity bound are in Appendix~A; the guarantee presupposes that $B_k$ attains its nominal coverage. Writing the decision margin as $\Delta_k=|G_k-\tau|$, whether the premise of Proposition~\ref{prop:sign-consistency} can be met is governed jointly by $J$, the calibration sample size, and $\Delta_k$. As $J\to0$ or $\Delta_k\to0$, the required evidence grows rapidly, and near-zero $J$ can defeat even a large margin because the posterior itself cannot be located reliably. In the experiments, VRR-Stop acts directly on the sign of $\hat G_k$; Eq.~\eqref{eq:sign-identifiability} and Proposition~\ref{prop:sign-consistency} serve as reliability-analysis tools for the flip-risk evaluation in the experiments.

\subsection{Guarded Keep-Best Fallback}
\label{ssec:guarded-fallback}

When a separation test on held-out labeled samples reports $\hat J$ near zero, or Eq.~\eqref{eq:sign-identifiability} fails throughout the evaluation window, calibrated stopping is no longer trustworthy and VRR-Guard should be used instead. The test compares acceptance rates of valid and invalid plans directly on labeled samples and does not depend on the binomial-mixture EM, avoiding diagnosing a possibly broken estimator with its own output. We partition evaluation scenarios offline by this criterion rather than deploying an online threshold switch.

VRR-Guard maintains the incumbent best candidate $c_k$, initialized as $c_0=p_0$. For any plan $p$, let $S(p)=\sum_{m=1}^{M}a^{(m)}(p)$ be its acceptance count, so $S_k=S(p_k)$. Given a retention margin $\delta$, the retention rule is given in Eq.~\eqref{eq:guarded-keep-best}:
\begin{equation}
\label{eq:guarded-keep-best}
c_k=
\begin{cases}
p_k, & S(p_k)\ge S(c_{k-1})+\delta,\\
c_{k-1}, & \text{otherwise},
\end{cases}
\end{equation}
and the incumbent $c_k$, not the last plan, is committed at termination.

\begin{lemma}[Bound on erroneous replacement]
\label{lem:guard-replacement}
Suppose the incumbent $c$ is truly valid and the new plan $p$ truly invalid, queries are conditionally independent given true states, and $J\ge0$. Then the per-round erroneous-replacement probability satisfies $\Pr[S(p)\ge S(c)+\delta]\le\exp\!\big(-(MJ+\delta)^2/(2M)\big)$, and the probability of at least one such replacement within $K_{\max}$ rounds is at most $K_{\max}$ times this bound.
\end{lemma}

The lemma follows from a Hoeffding bound on the difference of two independent binomials (proof and exact per-verifier tail probabilities in Appendix~A), showing that erroneous replacement decays exponentially in $\delta$ and $MJ$ and giving the margin an analytic basis. The bound is conservative; with $M=8$, the replay ablation of Appendix~I selects $\delta=5$ as the most robust setting. Because Eq.~\eqref{eq:guarded-keep-best} compares vote counts across rounds, it implicitly assumes within-trajectory stationarity of verifier noise; degradation under drift is tested in the shift experiments. VRR-Guard does not strictly dominate no-repair. It forfeits part of the achievable gain in repair-beneficial regimes in exchange for near-no-repair downside protection when calibration is unreliable.

Pseudocode integrating both stopping modes appears in Appendix~A, where the mode-selection threshold $J_{\min}$ corresponds to the held-out separation test above; scenarios are partitioned offline without tuning $J_{\min}$ online. Each round costs $M$ verifier calls plus a constant-time belief update, and repair calls occur only on \textsc{Repair}. Calibration is a one-off cost using repeated verification records and at most 300 labeled transitions. Hyperparameters are the verification budget $M$, gain threshold $\tau$, retention margin $\delta$, and repair budget $K_{\max}$.

\section{Experiments}
\label{sec:exp}

We organize the evaluation around four research questions.
\textbf{RQ1:} Does multi-round verify-repair monotonically improve true validity, or does reported acceptance rise while true validity falls?
\textbf{RQ2:} When repair can be harmful, does VRR-Stop avoid the degradation of heuristic stopping and approach the true-parameter myopic reference?
\textbf{RQ3:} With limited calibration samples and varying verifier discrimination, when is the stopping sign identifiable, and how does estimation error flip the decision?
\textbf{RQ4:} When calibration fails or the distribution shifts, does the conservative fallback retreat to near no-repair instead of collapsing?

\subsection{Setup}
\label{ssec:setup}

\textbf{Tasks, models, and protocol.}
The evaluation covers mathematical reasoning (GSM8K \citep{cobbe2021training}, MATH-500 \citep{hendrycks2021measuring,lightman2023lets}), code generation (MBPP \citep{austin2021program}), and tool use (BFCL \citep{patil2025bfcl}); true validity is judged by answer matching, symbolic verification, unit tests, and an executor, respectively. Generators and verifiers span the Qwen2.5 \citep{qwen2024qwen25}, Mistral \citep{jiang2023mistral7b}, and Llama \citep{grattafiori2024llama3} families. End-to-end evaluation uses $K_{\max}=5$ repair rounds and $M=8$ verification queries per round; the non-stationary loop-dynamics diagnostic runs to round 6. All calibration parameters are estimated with five-fold cross-fitting, so no instance's stopping decision touches its own label; the full protocol and model configurations are in Appendix~B.

\textbf{Metrics and baselines.}
The primary metric is the true validity $V$ of the final commit (95\% bootstrap confidence intervals from $B{=}10{,}000$ resamples) and the mean repair rounds $\langle K\rangle$; verifier reliability is described by $\rho_0,\rho_1$ and Youden's $J=1-\rho_0-\rho_1$; paired differences use exact McNemar tests and instance-level bootstrap. Baselines fall into three groups. Fixed-budget repair corresponds to the common configurations of Self-Refine, Reflexion, CRITIC, and SCoRe \citep{madaan2023selfrefine,shinn2023reflexion,gou2024critic,kumar2024training}. Heuristic stopping relies only on verification signals and includes majority stopping and confidence-threshold stopping \citep{wang2022selfconsistency}. Reference policies include no-repair, the true-parameter myopic (TPM) reference, and hindsight best-round selection, the latter two serving only as diagnostics. Formal definitions are in Appendix~A.

\subsection{Loop Dynamics}
\label{ssec:loop-dynamics}

\begin{figure}[t]
    \centering
    \includegraphics[width=0.87\linewidth]{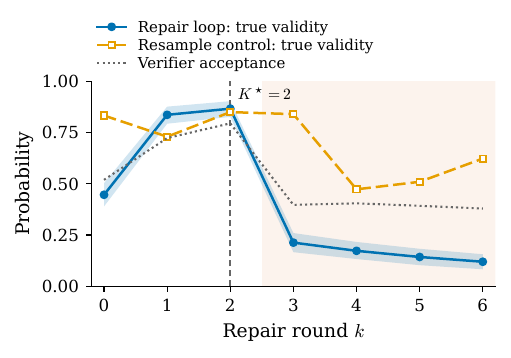}
    \caption{
    Loop dynamics under the non-stationary stress setting (\(N=300\), prompt mismatch from round \(3\)).
    True validity peaks at \(K^\star=2\) and then collapses, while independent resampling shows no comparable decline.
    }
    \label{fig:loop-dynamics}
\end{figure}

Addressing RQ1, Fig.~\ref{fig:loop-dynamics} shows that, under a $256$-token initial budget, validity starts at $0.45$, rises to $0.87$ after two repair rounds, then collapses after prompt mismatch is injected at round 3, reaching only $0.12$ at round 6, which forms an interior optimum at $K^\star=2$, whereas an independent-resampling control with the same budget stays between $0.47$ and $0.85$ throughout. The paired gain of the peak round over the final round is $74.7$ percentage points (95\% CI $[+69.3,+79.7]$), and the control exhibits no comparable path-dependent decline, so the degradation stems from repeated rewriting of the same candidate rather than sampling noise. Non-stationary loops thus contain an intrinsic boundary where the gain turns negative; reliable stopping depends on identifying the gain sign in time, not on preset budgets. The pattern is not specific to this diagnostic. Across eight settings, six exhibit monotone decline with damage probabilities between $0.615$ and $0.938$, BFCL multi-turn stays flat under a near-inert repair operator, and only the favorable setting improves, so harmful repair is the rule rather than the exception under stress. A stronger verifier does not remove the risk. With a process-reward-model verifier of $J=0.805$ on GSM8K, fixed five-round repair still drives validity from $0.727$ to $0.097$, and on MATH-500 ($J=0.77$) from $0.798$ to $0.150$ while the TPM reference holds $0.798$, so verifier quality strengthens stopping reliability but cannot eliminate repair damage (Appendix~H). On the $N=500$ stress traces, $55\%$ of instances see a correct plan repaired into an incorrect one, and $24\%$ of these damaging repairs win majority acceptance, raw-label statistics that are independent of any stopping rule (Appendix~C).

\subsection{Stopping Performance}
\label{ssec:stopping}

For RQ2, Table~\ref{tab:stopping-performance-main} evaluates end-to-end gains on the GSM8K / Qwen2.5-3B stress setting. Fixed five-round repair degrades the no-repair baseline from $0.700$ to $0.116$, whereas VRR-Stop reaches $0.722$ with only $0.72$ repair rounds on average. Against fixed five-round repair, the paired gain is $60.6$ percentage points (95\% CI $[+56.0,+65.0]$, McNemar $p<2\times10^{-85}$); VRR-Stop also beats majority stopping ($+3.2$ pp, $p<2\times10^{-4}$) and the TPM reference ($+2.8$ pp, $p<6\times10^{-4}$), while its difference from no-repair ($+2.2$ pp) has a CI crossing zero. Under the fixed-budget deployment mode of iterative-feedback baselines, Reflexion and Self-Refine end at $0.095$ and $0.080$ on this stress setting, while VRR-Stop on the same batches of trajectories reaches $0.740$ and $0.710$. The ordering replicates across sample sizes and domains, with three disjoint GSM8K windows giving $0.752\pm0.037$ against the reference $0.757\pm0.058$ and Qwen-7B stress retaining $0.875$ by learning to never repair. Calibrated stopping thus terminates the loop in time when repair is harmful, preserving validity at lower cost than heuristic voting; all remaining baselines and paired tests are in Appendices~D and~G.

\begin{table}[t]
\centering
\small
\caption{Stopping performance on the GSM8K / Qwen2.5-3B prompt-mismatch stress setting (\(N{=}500\), \(M{=}8\), \(K_{\max}{=}5\)), abbreviated; the TPM reference applies the same stopping rule with ground-truth parameters as a diagnostic.}
\label{tab:stopping-performance-main}
\begin{tabular}{lcc}
\toprule
Method & True validity \(V\) [95\% CI] & \(\langle K\rangle\) \\
\midrule
No repair                & 0.700 [.658,.740] & 0.00 \\
Majority stopping        & 0.690 [.648,.730] & 0.92 \\
ConfStop-0.85            & 0.562 [.518,.604] & 1.92 \\
Fixed repair \(K{=}5\)   & 0.116 [.088,.144] & 5.00 \\
VRR-Stop                 & \textbf{0.722} [.682,.760] & 0.72 \\
\midrule
TPM reference            & 0.694 [.652,.734] & 0.89 \\
\bottomrule
\end{tabular}
\end{table}

\subsection{Calibration and Identifiability}
\label{ssec:calibration-identifiability}

\begin{figure}[t]
    \centering
    \includegraphics[width=0.87\linewidth]{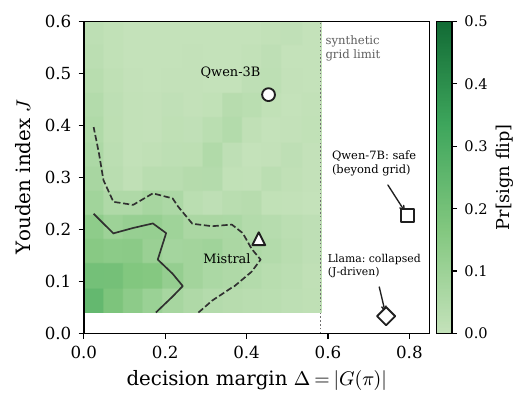}
    \caption{
Stopping-sign flip probability over \(J\) and \(\Delta\) under the deployed rule; markers denote real verifiers.
Flip risk concentrates at low \(J\) and low \(\Delta\), and near-zero \(J\) fails even with a large margin (Llama).
}
    \label{fig:identifiability}
\end{figure}

For RQ3, we sweep a controlled $J$--$\Delta$ grid ($\Delta$ is the decision margin of the calibration subsection) and evaluate four cross-family verifiers on GSM8K, with $N=300$ calibration samples and $M=8$ queries per plan. Figure~\ref{fig:identifiability} shows that flip risk concentrates where low $J$ meets low $\Delta$. The flip probability reaches $0.183$ for $J\le0.15$ and $\Delta\le0.10$, versus $0.014$ for $J\ge0.4$ or $\Delta\ge0.30$, a roughly $13$-fold gap. Among the real verifiers, calibrated stopping stays within $2.8$ percentage points of the TPM reference for Qwen-3B, Qwen-7B, and Mistral (all $J\ge0.18$), whereas Llama has $J$ of only $0.03$. Despite a decision margin of about $0.74$, its validity collapses from the reference $0.803$ to $0.223$ ($-58.0$ pp, 95\% CI $[-63.7,-52.0]$, McNemar $p<2\times10^{-48}$). The failure mechanism is that the likelihood surface of the binomial-mixture EM flattens as $J\to0$. With $N=120$ the estimate is $\hat\rho_1=0.27$, yet enlarging the sample to $300$ worsens it to $0.077$ against a true $0.609$, and a parameter sweep shows calibrated stopping recovers only for $\hat\rho_1\gtrsim0.30$. Stopping reliability is thus determined by whether finite evidence can stabilize the gain sign, not by the absolute size of parameter error; near-zero $J$ mandates abandoning fine-grained calibration for the conservative fallback (Appendix~E).

\subsection{Guarded Fallback under Shift}
\label{ssec:guarded-shift}

For RQ4, Table~\ref{tab:robustness-main} examines calibration failure and distribution shift. Fixed five-round repair falls far below no-repair in most shifted settings (e.g., Qwen-7B drops from $0.875$ to $0.075$), whereas VRR-Guard stays near the no-repair level in all seven settings and never collapses. In the Llama setting where calibrated stopping fails, VRR-Guard recovers to $0.793$, exceeding the failed VRR-Stop by $57.0$ percentage points (95\% CI $[+51.0,+62.7]$, McNemar $p<1\times10^{-47}$) and fixed five-round repair by $73.7$ points. Its conservatism costs about $2.0$ and $0.3$ pp against no-repair on Mistral and BFCL (the latter's paired CI contains zero), and its $0.810$ in the favorable setting trails fixed five-round repair's $0.875$. Against margin-free verifier-best selection, the retention margin contributes $+45.0$ pp in the Llama scenario and between $-4.0$ and $+8.0$ pp elsewhere, concentrating its value where verification signals are least trustworthy. The value of VRR-Guard is to pull the system back to a near-no-repair safe state when the sign is unidentifiable, not to dominate everywhere; full results and paired tests are in Appendices~F and~G. Ablations of the calibration input, the retention margin $\delta$, and the calibration sample size corroborate these design choices, with binomial EM the best input, $\delta=5$ most robust, and $N_{\text{calib}}=300$ sufficient (Appendix~I).

\begin{table}[t]
\centering
\small
\setlength{\tabcolsep}{3pt}
\caption{Robustness under calibration failure and distribution shift, replayed on saved trajectories with \(M=8\) and \(\delta=5\); shift codes M, V, T denote model, verifier, and task shift.}
\label{tab:robustness-main}
\resizebox{\columnwidth}{!}{
\begin{tabular}{lccccc}
\toprule
Setting & \(J\) & None & Fixed 5 & Calib. & Guard \\
\midrule
Qwen-3B fav.\ (--)   & 0.39 & 0.740 & 0.875 & 0.845 & 0.810 \\
Qwen-3B stress (--)  & 0.46 & 0.700 & 0.116 & \textbf{0.722} & \textbf{0.742} \\
Qwen-7B stress (M)   & 0.23 & 0.875 & 0.075 & 0.875 & 0.875 \\
Mistral-7B stress (V)& 0.18 & 0.507 & 0.047 & 0.463 & 0.487 \\
Llama-3-8B stress (M)& \textbf{0.03} & 0.803 & 0.057 & 0.223 & \textbf{0.793} \\
MATH-500 (T)         & 0.22 & 0.798 & 0.150 & 0.798 & 0.796 \\
BFCL single (T)      & 0.07 & 0.812 & 0.372 & 0.782 & 0.810 \\
\bottomrule
\end{tabular}
}
\end{table}

\section{Conclusion}
\label{sec:conclusion}

This paper addresses the stopping decision in LLM verify-repair loops, whose crux is that the paradigm implicitly assumes multi-round repair improves true quality and verification signals represent true validity, and both assumptions fail together when the verifier and the repairer are noisy. We propose VRR-Stop, which builds on a four-parameter noise model and couples belief filtering with a marginal-gain sign criterion, turning stopping into a calibratable, identifiable sign decision, with the fallback VRR-Guard covering the unidentifiable regime. On the GSM8K / Qwen2.5-3B stress setting, VRR-Stop improves final true validity by $60.6$ percentage points over fixed five-round repair and slightly exceeds the true-parameter myopic reference, while VRR-Guard restores validity from $0.223$ to $0.793$ when calibration fails at near-zero discrimination. The method is limited by its local-stationarity approximation and binary validity representation; future work will model round-varying repair dynamics and online switching between the two modes.

\section*{Acknowledgments}
This work was supported in part by the Brain Science and Brain-like Intelligence Technology-National Science and Technology Major Project (No. 2021ZD0200600, No. 2021ZD0200408), and in part by the National Natural Science Foundation of China (Grant No. U24B20186), and supported by the Supercomputing Center of Lanzhou University.

\appendix

\section{Derivations, Algorithms, and Reference Policies}
\label{app:derivations}

This appendix provides the complete derivations of the stopping quantities in the main text, proofs of the two formal results, the construction of the calibration estimators and the error radius, and formal definitions of all comparison policies. Pseudocode integrating the two stopping modes is given in Algorithm~\ref{alg:vrr}; the mode-selection threshold $J_{\min}$ corresponds to the separability check on held-out labeled samples in the guarded-fallback subsection of the main text. Evaluation scenarios were partitioned offline according to this check, and the specific value of $J_{\min}$ was not tuned online.

\subsection{Belief Recursion and the Stopping Boundary}

\begin{figure}[t]
\centering
\includegraphics[width=\columnwidth]{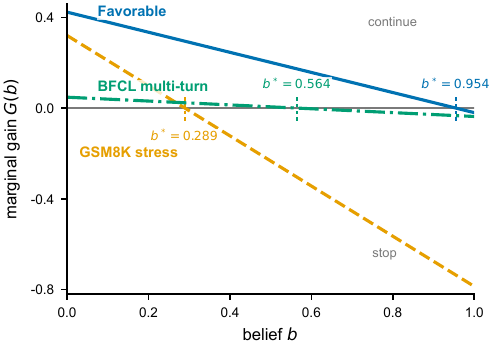}
\caption{Expected benefit $G(b)=(1-b)\alpha-b\beta$ of running one more repair round, where $b$ is the probability that the current plan is already correct, $\alpha$ is the probability that repair fixes an incorrect plan, and $\beta$ is the probability that repair breaks a correct one; the three lines use the $(\alpha,\beta)$ values measured in Table~\ref{tab:loop-dynamics}. Wherever a line lies above zero, one more repair is expected to help; below zero it is expected to hurt; the crossing $b^\star=\alpha/(\alpha+\beta)$ (vertical dashes) marks the confidence level at which repairing stops paying off. Because $b^\star$ ranges from $0.954$ (favorable) down to $0.289$ (stress), the same confidence can mean continue in one setting and stop in another, so no fixed round budget or universal confidence threshold is safe across settings.}
\label{fig:gain-geometry}
\end{figure}

Conditioned on the true state $y_k$, the $M$ verification queries are conditionally independent, so the acceptance count is binomial: $S_k\mid y_k{=}1\sim\mathrm{Bin}(M,1-\rho_1)$ and $S_k\mid y_k{=}0\sim\mathrm{Bin}(M,\rho_0)$. Applying Bayes' rule to the predictive belief $b_k^{-}$, the two binomial coefficients cancel between numerator and denominator, which yields Eq.~\eqref{eq:belief-update}. Given $y_k$, the repair transition is conditionally independent of the history $H_k$, so $\Pr(y_{k+1}{=}1\mid H_k,\mathrm{repair})=(1-\beta)b_k+\alpha(1-b_k)$; this is the prediction step. Subtracting the current belief from the predictive belief gives Eq.~\eqref{eq:gated-gain}. Setting $G_k=0$ yields the boundary $b^\star=\alpha/(\alpha+\beta)$ of Eq.~\eqref{eq:gated-boundary}, which is also the fixed point of the affine map $b\mapsto(1-\beta)b+\alpha(1-b)$ and coincides with the fixed point of the population dynamics in the motivation subsection of the main text. Under a nonzero threshold, $G_k>\tau$ is equivalent to $b_k<b_\tau^\star=(\alpha-\tau)/(\alpha+\beta)$. Fig.~\ref{fig:gain-geometry} illustrates the geometry of this boundary with three empirically measured repair operators.

\subsection{Proof of Proposition 1}

Let $E=\{|\hat G_k-G_k|\le B_k\}$; by the definition of $B_k$, $\Pr(E)\ge1-\eta$. If $\hat G_k-B_k>\tau$, then on $E$ we have $G_k\ge\hat G_k-B_k>\tau$, so the true action and the estimated action are both \textsc{Repair}. If $\hat G_k+B_k\le\tau$, then on $E$ we have $G_k\le\hat G_k+B_k\le\tau$, so both are \textsc{Commit}. In either case the two actions agree with probability at least $\Pr(E)\ge1-\eta$, which completes the proof.

\subsection{Proof and Numerical Evaluation of Lemma 1}

Suppose the current candidate $c$ is truly valid and the new plan $p$ is truly invalid; then $S(p)\sim\mathrm{Bin}(M,\rho_0)$ and $S(c)\sim\mathrm{Bin}(M,1-\rho_1)$, and the two are independent given the true states. Pairing the $m$-th queries, the differences $X_m=a^{(m)}(p)-a^{(m)}(c)\in[-1,1]$ are i.i.d.\ with $\mathbb{E}[X_m]=\rho_0-(1-\rho_1)=-J$. Applying Hoeffding's inequality to $D=\sum_{m=1}^{M}X_m$ gives $\Pr(D-\mathbb{E}[D]\ge t)\le\exp(-t^{2}/(2M))$; taking $t=MJ+\delta\ge0$ yields the single-round bound, and the statement over $K_{\max}$ rounds follows from a union bound, which completes the proof.

This bound is conservative relative to the exact tail probability. The exact value is the finite sum $\Pr[S(p)-S(c)\ge\delta]=\sum_{i-j\ge\delta}\Pr[S(p){=}i]\,\Pr[S(c){=}j]$. Substituting the calibrated values of each verifier in Table~\ref{tab:loop-dynamics} ($M=8$, $\delta=5$), the single-round erroneous-replacement probabilities are $5.6\times10^{-6}$ for Qwen-3B, $1.0\times10^{-6}$ for Qwen-7B, $5.3\times10^{-5}$ for Mistral, and $5.9\times10^{-3}$ for Llama; even in the worst case of near-zero $J$, the five-round union bound does not exceed $3\%$.

\subsection{Calibration Estimators and Identifiability}

Calibration from repeated verification records treats the round-1 acceptance counts as a two-component binomial mixture, $S\sim\pi\,\mathrm{Bin}(M,1-\rho_1)+(1-\pi)\,\mathrm{Bin}(M,\rho_0)$. The E-step of expectation-maximization (EM) computes each plan's responsibility of belonging to the valid component, and the M-step updates the mixture weight and the two acceptance rates from responsibility-weighted frequencies; after multiple random restarts the solution with the highest log-likelihood is retained, and the component labels are fixed by the convention that the component with the higher acceptance rate corresponds to the valid class. The mixture is identifiable if and only if the two component acceptance rates differ, that is, $J\ne0$. As $J\to0$ the likelihood surface flattens and the estimator no longer carries information about $\rho_0,\rho_1$; this is exactly the mechanism behind the failure of calibrated stopping in the Llama scenario. The transition parameters are estimated from labeled pre/post-repair plan pairs: $\hat\alpha$ is the frequency of invalid-to-valid transitions and $\hat\beta$ the frequency of valid-to-invalid transitions. The prior $\hat\pi$ is the validity rate of initial plans on the calibration folds, and all estimation is performed outside the held-out folds of the cross-fitting.

\subsection{Error Radius and Sample Complexity}

The error radius $B_k$ is constructed by propagating confidence bounds on the calibrated estimates through Eq.~\eqref{eq:belief-update} and the expression for $\hat G_k$, taking the worst-case excursion of $\hat G_k$ over the endpoints of each parameter interval. In an idealized single-query loop, the calibration sample size required for the stopping sign to be identifiable satisfies $N\gtrsim C_1\log(1/\delta_c)/(M J^2\Delta^2)$, where $\delta_c$ is the allowed failure probability. This bound is derived under the idealized single-query loop; we do not claim a corresponding explicit bound for the deployed rule, and decision reliability under deployment conditions is characterized directly by the flip-risk simulation in the identifiability experiments of the main text.

\subsection{Formal Definitions of Baselines and References}

No-repair commits $p_0$ directly. Fixed-$K$ repairs unconditionally for $K$ rounds and then commits $p_K$. Majority stopping commits the current plan when $S_k>M/2$ and otherwise keeps repairing until the budget is exhausted. ConfStop-$\tau_{\text{conf}}$ replaces the commit condition with $S_k/M\ge\tau_{\text{conf}}$. Accepted-first commits the first candidate on the trajectory that receives majority acceptance, and commits the final round at budget exhaustion when no round is majority-accepted. Last-accepted commits the last majority-accepted candidate, and commits the initial plan when no round is accepted. Verifier-best-of-trajectory commits the candidate with the highest acceptance count, breaking ties toward the earlier round.
The true-parameter myopic (TPM) reference executes the same myopic stopping rule as VRR-Stop but with the true loop parameters $(\rho_0,\rho_1,\alpha,\beta,\pi)$; it isolates the effect of calibration error on the decisions and is not a hindsight policy that selects trajectories after observing ground truth. Hindsight best-round selection picks the best round on each trajectory according to the true labels; it serves only as a trajectory-structure diagnostic and is not deployable. The stationarity and independence assumptions of the four-parameter model are local approximations; their scope of validity and failure modes are discussed in Appendix~\ref{app:extended-discussion}.

\begin{algorithm}[h]
\caption{VRR-Stop with the guarded fallback (intended deployment procedure; experiments evaluate the two modes separately on saved trajectories)}
\label{alg:vrr}
\begin{algorithmic}[1]
\REQUIRE plan $p_0$; calibrated $\hat\pi,\hat\rho_0,\hat\rho_1,\hat\alpha,\hat\beta$; thresholds $\tau,\eta,J_{\min}$; margin $\delta$; budgets $M,K_{\max}$
\STATE $\mathrm{mode}\leftarrow\textsc{Stop}$ \textbf{if} labeled-sample diagnostic gives $\hat J\ge J_{\min}$ \textbf{else} $\textsc{Guard}$
\STATE $b^{-}\leftarrow\hat\pi$;\quad $c\leftarrow p_0$;\quad $S_c\leftarrow-\infty$
\FOR{$k=0,1,\ldots,K_{\max}$}
\STATE query the verifier $M$ times on $p_k$ and collect $S_k$
\STATE $b_k\leftarrow$ posterior update of $b^{-}$ with $S_k$ \COMMENT{Eq.~\eqref{eq:belief-update}}
\IF{$S_k\ge S_c+\delta$}
\STATE $c\leftarrow p_k$;\quad $S_c\leftarrow S_k$ \COMMENT{keep-best tracking, Eq.~\eqref{eq:guarded-keep-best}}
\ENDIF
\IF{$\mathrm{mode}=\textsc{Stop}$}
\STATE $\hat G_k\leftarrow(1-b_k)\hat\alpha-b_k\hat\beta$;\quad construct $B_k$
\IF{$\hat G_k+B_k\le\tau$}
\RETURN $p_k$ \COMMENT{commit, stop sign identifiable}
\ELSIF{$\hat G_k-B_k\le\tau$}
\STATE $\mathrm{mode}\leftarrow\textsc{Guard}$ \COMMENT{sign not identifiable, Eq.~\eqref{eq:sign-identifiability}}
\ENDIF
\ENDIF
\IF{$k=K_{\max}$}
\RETURN $p_k$ \textbf{if} $\mathrm{mode}=\textsc{Stop}$ \textbf{else} $c$
\ENDIF
\STATE $p_{k+1}\leftarrow$ repair $p_k$ with verifier feedback
\STATE $b^{-}\leftarrow(1-\hat\beta)b_k+\hat\alpha(1-b_k)$ \COMMENT{predict step}
\ENDFOR
\end{algorithmic}
\end{algorithm}

\section{Full Experimental Setup}
\label{app:full-setup}

This appendix provides the complete experimental setup required for auditing and reproduction; the runtime environment, random seeds, and prompt templates are given in Appendix~\ref{app:reproducibility}.

\textbf{Task windows and verifiers.} Table~\ref{tab:app-datasets} lists the verifier type, evaluation windows, and label source of each task; the verifier for MATH-500 is a process reward model (PRM). The calibration probe issues \(16\) verification queries per plan to stabilize the acceptance-rate estimates, whereas the end-to-end loop issues \(M=8\) queries per round; the two share the same verifier configuration.
The calibration stage uses at most \(300\) labeled samples.

\textbf{Construction of evaluation settings.} The favorable setting is the standard generate-verify-repair loop without perturbation. In the prompt-mismatch (PM) stress setting, the copy of the problem statement received by the repairer is injected with numeric- or condition-level perturbations, while the verifier and the ground-truth judgment always use the original statement; the repairer can therefore produce plausible-looking but incorrect revisions from a corrupted premise. The repair temperature is \(1.0\) in the stress setting and \(0.7\) elsewhere. The main text's non-stationarity diagnostic adds two further constructions: initial drafts are generated under a \(256\)-token budget, and perturbations are injected from round \(3\) onward and run through round \(6\).

\textbf{Calibration protocol and cross-fitting.} To prevent the same instance from being used both for parameter estimation and for stopping evaluation, the VRR-Stop results and the true-parameter myopic (TPM) reference reported in the main text both use five-fold cross-fitting: the stopping decision for each instance uses only the \(\rho_0,\rho_1,\alpha,\beta\) and prior parameters estimated from the other four folds. Decisions from the five held-out folds are pooled at the instance level before computing the final validity and the mean number of repair rounds; the random seed of the fold split is fixed and released with the replay artifacts. In-sample calibration results serve only as a sensitivity audit, not as the deployment protocol. The prior \(\hat\pi\) is the frequency estimate of the true validity rate of initial plans on the calibration folds. The transition parameters are estimated from the pre/post plan pairs of the first repair round on the frozen trajectories; during data generation, repair was triggered in every round without verifier gating, so the transition samples are free of selection bias induced by the acceptance signal.

\textbf{Definition of shift scenarios.} Each shift scenario reruns the entire loop (generation, repair, and judging) on the target model or task. The shift codes in Table~\ref{tab:robustness-main} mark the dominant change dimension of each scenario relative to the Qwen-3B baseline: M denotes a change of generator family or scale, V a change of verifier family, and T a change of task together with its accompanying verifier; the evaluation protocol and the calibration pipeline remain unchanged. The generator and verifier models of each setting are listed in Table~\ref{tab:app-setting-models}.

\begin{table}[h]
\centering
\small
\setlength{\tabcolsep}{3pt}
\caption{Generator and verifier models of each evaluation setting. In
every shift scenario the whole loop (generation, repair, and judging) runs
on the target models; the shift codes of Table~\ref{tab:robustness-main}
mark the dominant change dimension relative to the Qwen-3B baseline.}
\label{tab:app-setting-models}
\resizebox{\columnwidth}{!}{
\begin{tabular}{lll}
\toprule
Setting & Generator & Verifier \\
\midrule
Qwen-3B fav./stress (--) & Qwen2.5-3B-Instruct & same model (LLM judge) \\
Qwen-7B stress (M)       & Qwen2.5-7B-Instruct & same model (LLM judge) \\
Mistral-7B stress (V)    & Mistral-7B-Instruct-v0.3 & same model (LLM judge) \\
Llama-3-8B stress (M)    & Llama-3-8B-Instruct & same model (LLM judge) \\
MATH-500 (T)             & Qwen2.5-Math-7B-Instruct & Qwen2.5-Math-PRM-7B \\
BFCL single (T)          & Qwen2.5-7B-Instruct & same model (LLM judge) \\
BFCL multi (T)           & Qwen2.5-7B-Instruct & official executor \\
\bottomrule
\end{tabular}
}
\end{table}

\textbf{Hyperparameters.} End-to-end evaluation uniformly uses \(M=8\), \(K_{\max}=5\) (the non-stationarity diagnostic runs through round \(6\)), \(\tau=0\), \(\delta=5\), and \(B=10{,}000\) bootstrap resamples. The coverage level \(\eta\) is a confidence parameter of the reliability-analysis layer and does not enter end-to-end deployment. Only two hyperparameters were tuned: the retention margin \(\delta\) was swept over \(\{0,\ldots,8\}\) by pure replay on all seven settings, selecting the value most robust across settings, and the calibration sample size was examined over \(N_{\text{calib}}\in\{50,100,200,300\}\). The remaining settings were fixed a priori and not tuned: \(M=8\) and \(K_{\max}=5\) follow common verify-repair budgets, \(\tau=0\) by definition of the cost-free objective, and systematic sweeps of \(M\) and \(\tau\) are left to future work (Appendix~\ref{app:ablations}).
A manifest mapping each figure and table of the main text to its raw data artifacts is released with the anonymous code repository.

\begin{table}[h]
\centering
\small
\setlength{\tabcolsep}{3pt}
\caption{Datasets, verifiers, and evaluation windows. Probe windows feed
the calibration probe (16 votes per plan); e2e and stress windows feed
end-to-end evaluation (\(M{=}8\)).}
\label{tab:app-datasets}
\begin{tabular}{lll}
\toprule
Task & Verifier type & Windows (\(N\)) \\
\midrule
GSM8K    & LLM judge   & \begin{tabular}[t]{@{}l@{}}probe 300 / e2e 200\\ stress 200, 500\end{tabular} \\
MATH-500 & PRM         & 500 \\
MBPP     & LLM judge   & 150 \\
BFCL     & LLM judge / executor & 400 / 199 \\
\bottomrule
\end{tabular}
\end{table}

\section{Full Loop Dynamics and Failure Cases}
\label{app:full-dynamics}

This appendix examines whether the phenomenon of RQ1 persists across settings and unfolds the complete chain of a harmful repair on a single instance.

\subsection{Cross-Setting Loop Dynamics}

\begin{figure}[t]
\centering
\includegraphics[width=\columnwidth]{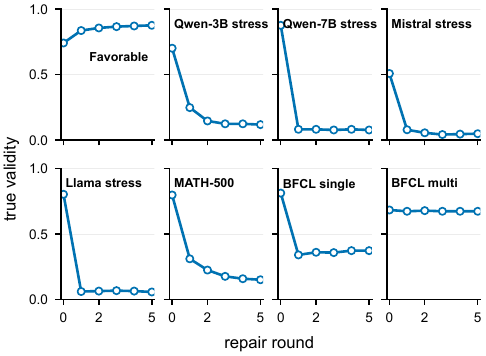}
\caption{Mean true validity per repair round when every plan is repaired unconditionally, averaged over each trace's evaluation window; the eight panels are the settings of Table~\ref{tab:loop-dynamics}. Round $0$ equals the no-repair baseline and round $k$ equals the fixed-budget value Fixed $K{=}k$, so each curve traces the full path between the two extremes reported in the tables. Six settings collapse within the first rounds and never recover, BFCL multi-turn stays flat, and only the favorable setting improves.}
\label{fig:cross-setting-dynamics}
\end{figure}

Table~\ref{tab:loop-dynamics} summarizes the round-1 parameter estimates and trajectory shapes of the eight settings on their respective evaluation windows. Apart from the favorable setting, which rises monotonically, and BFCL multi-turn, which is approximately flat, true validity declines monotonically with the round index in the remaining six settings; the damage probability $\beta$ ranges from $0.615$ to $0.938$ and is typically several times the repair probability $\alpha$, so the stopping boundary $b^\star$ is pushed below $0.29$. The MATH-500 row shows that a strong verifier ($J=0.77$) does not prevent a high-$\beta$ repairer from damaging correct plans, whereas the near-inert operator of BFCL multi-turn ($\alpha=0.02$, $\beta=0.04$) provides a counterexample of multi-round repair without collapse. The finite-trace diagnostic in the last column shows that, outside the flat setting, the gap between myopic stopping and the hindsight best round is at most $4$ percentage points. Fig.~\ref{fig:cross-setting-dynamics} plots the per-round true-validity trajectories of all settings.

\begin{table*}[t]
\centering
\small
\setlength{\tabcolsep}{4pt}
\caption{Full loop dynamics across the eight settings, where PM-stress
denotes the prompt-mismatch stress setting. All parameters
are per-call round-1 estimates on each trace's evaluation window (the
Qwen2.5-3B PM-stress row uses the \(N{=}500\) window),
\(b^\star=\alpha/(\alpha+\beta)\) is the stopping boundary of
Eq.~\eqref{eq:gated-boundary}, and the last column is a finite-trace
diagnostic rather than a formal dynamic-programming regret. Read each row
by weighing repair \(\alpha\) against damage \(\beta\): wherever \(\beta\)
dominates, \(b^\star\) collapses toward zero and the trajectory shape turns
into a monotone decline, so harmful repair is the rule across the stress
rows rather than an exception.}
\label{tab:loop-dynamics}
\begin{tabular}{llccccccccc}
\toprule
Setting & Model & Verifier & \(\rho_0\) & \(\rho_1\) & \(J\) & \(\alpha\) & \(\beta\) & \(b^\star\)~(Eq.~7) & Shape & Best-\(K\) \(-\) myopic \\
\midrule
Favorable & Qwen2.5-3B & LLM judge & 0.438 & 0.169 & 0.39 & 0.423 & 0.020 & 0.954 & monotone \(\uparrow\) & 0.025 \\
PM-stress & Qwen2.5-3B & LLM judge & 0.364 & 0.177 & 0.46 & 0.320 & 0.786 & 0.289 & monotone \(\downarrow\) & 0.016 \\
PM-stress & Qwen2.5-7B & LLM judge & 0.725 & 0.048 & 0.23 & 0.000 & 0.909 & 0.000 & monotone \(\downarrow\) & 0.000 \\
PM-stress & Mistral-7B & LLM judge & 0.707 & 0.111 & 0.18 & 0.014 & 0.862 & 0.015 & monotone \(\downarrow\) & 0.040 \\
PM-stress & Llama-3-8B & LLM judge & 0.358 & 0.609 & 0.03 & 0.051 & 0.938 & 0.051 & monotone \(\downarrow\) & 0.000 \\
MATH-500 & Qwen2.5-Math-7B & PRM & 0.109 & 0.118 & 0.77 & 0.020 & 0.617 & 0.031 & monotone \(\downarrow\) & 0.000 \\
BFCL single & Qwen2.5-7B & LLM judge & 0.875 & 0.057 & 0.07 & 0.147 & 0.615 & 0.192 & monotone \(\downarrow\) & 0.027 \\
BFCL multi & Qwen2.5-7B & Executor & 0.659 & 0.223 & 0.12 & 0.048 & 0.037 & 0.564 & flat & \(-0.015\) \\
\bottomrule
\end{tabular}
\end{table*}

\subsection{A Representative Failure Trace}

\begin{figure*}[t]
\centering
\includegraphics[width=\textwidth]{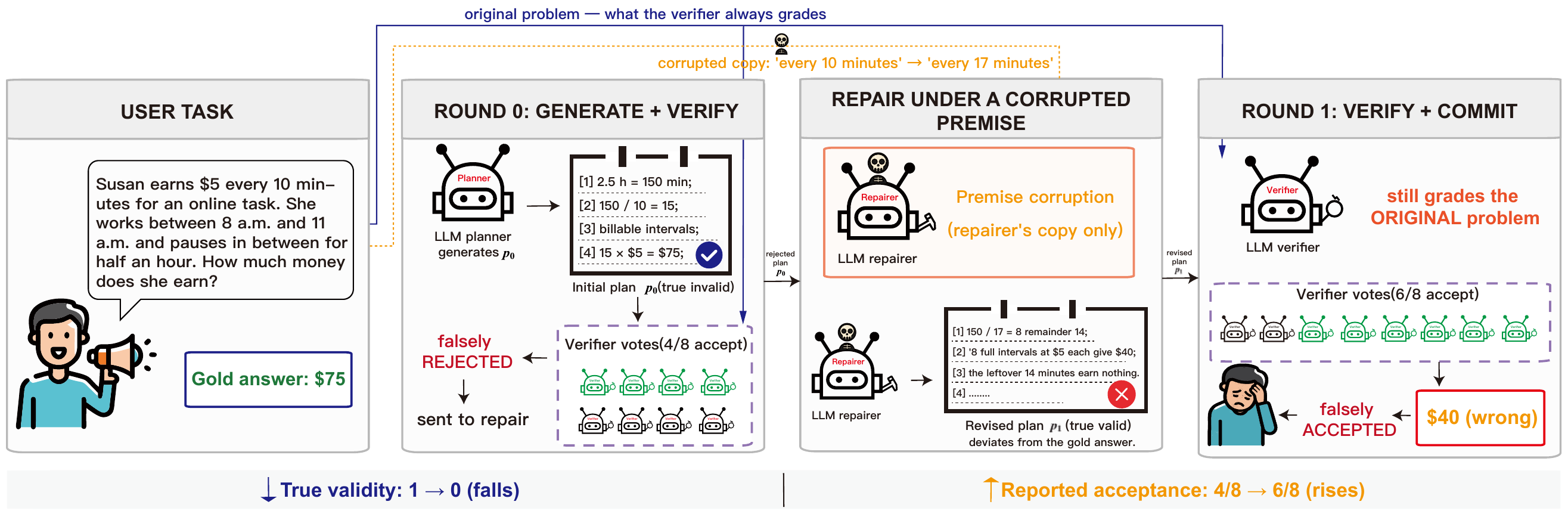}
\caption{End-to-end failure trace of GSM8K instance \#292. The two top rails
show the fork that drives it: the original problem (blue, solid) is what the
verifier always grades, while the corrupted copy (amber, dashed) that turns
every $10$ minutes into every $17$ minutes reaches only the repairer. A valid
plan under-accepted at $4/8$ is competently rewritten from the corrupted
premise into an invalid one, which the verifier accepts at $6/8$ and commits
to the user, so true validity falls from $1$ to $0$ while reported acceptance
rises.}
\label{fig:failure-trace}
\end{figure*}

Table~\ref{tab:loop-dynamics} and Fig.~\ref{fig:loop-dynamics} give the aggregate view; this section unfolds the same mechanism on a single instance. The pattern is not a hand-picked pathology: on the \(N=500\) stress traces, the round-1 validity of \(0.70\) drops to \(0.25\) after a single repair round, \(55\%\) of instances experience a correct plan being repaired into an incorrect one, and \(24\%\) of these damaging repairs receive majority acceptance among the eight judgments. All of these are raw-label statistics, independent of the stopping rule and its calibration. The trace comes from GSM8K test instance \#292 (Qwen2.5-3B, \(M=8\) judgments per round); the complete causal chain is shown in Fig.~\ref{fig:failure-trace}.

As shown in Fig.~\ref{fig:failure-trace}, the initial plan correctly derives \(75\) dollars from \(150/10=15\) billing intervals, yet receives only four of the eight votes and suffers a false rejection. The upstream perturbation corrupts only the repairer's copy of the problem statement, changing every \(10\) minutes to every \(17\) minutes; the repairer then competently rewrites the plan from the corrupted premise into an incorrect answer of \(40\) dollars, and the verifier, which judges against the uncorrupted original problem, instead gives it six votes and commits it. True validity flips from \(1\) to \(0\) while the reported acceptance rises. This is the trade-off between the repair benefit \((1-b_k)\alpha\) and the damage loss \(b_k\beta\) of the belief-filtering subsection of the main text unfolding on a single instance: a high-\(\beta\) rewrite compounded by the missed detection of a verifier false acceptance.

The case text is taken from a replay capture of the stress-setting pipeline, and a second instance exhibits the same type of chain; both full transcripts (including the eight judgment responses per round, the two candidate solutions, and the exact repair prompts) are released with the anonymous code repository. Per-instance sampling is not bitwise reproducible across inference-engine versions, so the case cites the replay transcripts rather than the original frozen trajectories; the aggregate conclusions are unaffected.

\section{Full Stopping and Cross-Benchmark Results}
\label{app:full-stopping}

This appendix examines the stability of the RQ2 conclusions across additional baselines, larger samples, different models, and different tasks; all numbers are reported under the five-fold cross-fitting protocol.

\textbf{Full baselines.} Table~\ref{tab:app-full-stopping} completes the full set of comparison strategies on the stress setting of Table~\ref{tab:stopping-performance-main}, where \(\Delta V\) denotes the difference relative to the true-parameter myopic (TPM) reference. The three families of baselines exhibit a clear stratification. Heuristic stopping rules that rely solely on the verifier signal approach but never exceed no-repair (Majority stopping and Accepted-first both at \(0.690\), Verifier-best at \(0.696\)); confidence-threshold stopping and Last-accepted degrade moderately (\(0.562\) and \(0.504\)); fixed-budget repair collapses severely and worsens monotonically with the budget (\(0.246\), \(0.122\), and \(0.116\) for \(K=1,3,5\)). VRR-Stop at \(0.722\) is the only deployable strategy that exceeds both the no-repair baseline and the TPM reference. The reference row at the bottom of the table is diagnostic only and is not deployable.

\textbf{Sample size and multi-window replication.} On the \(N=200\) window of the same setting, VRR-Stop attains \(0.705\) (95\% CI \([0.640,0.765]\), \(\langle K\rangle=0.82\)) against a TPM reference of \(0.695\), directionally consistent with \(0.722\) versus \(0.694\) at \(N=500\); the confidence intervals shrink by roughly a factor of \(\sqrt{2.5}\), and the strategy ranking is unchanged. Across three disjoint GSM8K windows, calibrated stopping attains \(0.752\pm0.037\), the TPM reference \(0.757\pm0.058\), and fixed \(5\)-round repair \(0.118\pm0.007\). Calibrated stopping and the reference strategy alternate in per-window ranking; the two are close but not identical, which is the expected behavior of a calibration that tracks the true parameters.

\textbf{Cross-model and cross-domain replication.} The Qwen2.5-7B stress setting exhibits the largest collapse in the paper: fixed \(5\)-round repair falls from the no-repair level of \(0.875\) to \(0.075\), whereas VRR-Stop preserves \(0.875\) with zero repair rounds on average, matching the TPM reference; in this setting, calibrated stopping learns not to repair at all. On MBPP, under both the favorable and the stress settings, calibrated stopping coincides with the TPM reference (\(0.533\) and \(0.493\), respectively), remaining stable across domains and calibration protocols.

\textbf{Direct comparison with iterative-feedback baselines.} Under the fixed-budget deployment mode of the stress setting, Reflexion and Self-Refine reach final validities of \(0.095\) and \(0.080\), whereas VRR-Stop on the same trajectories attains \(0.740\) and \(0.710\) (TPM reference: \(0.710\) and \(0.705\)). This comparison audits the fixed-budget deployment mode of iterative feedback and does not claim to reproduce the full training details of the original methods.

\begin{table*}[t]
\centering
\small
\caption{Full stopping results on the GSM8K / Qwen2.5-3B prompt-mismatch
stress setting (\(N{=}500\)); \(\Delta V\) is measured against the TPM
reference (positive = better), verifier-only baselines are offline replays
on the same saved trajectories, and the final row is a non-deployable
diagnostic. VRR-Stop is the only deployable strategy above both the
no-repair baseline and the TPM reference.}
\label{tab:app-full-stopping}
\begin{tabular}{lccc}
\toprule
Method & \(V\) [95\% CI] & \(\langle K\rangle\) & \(\Delta V\) (pp) \\
\midrule
No repair & 0.700 [.658,.740] & 0.00 & \(+0.6\) \\
Majority stopping & 0.690 [.648,.730] & 0.92 & \(-0.4\) \\
ConfStop-0.85 & 0.562 [.518,.604] & 1.92 & \(-13.2\) \\
Fixed repair \(K{=}1\) & 0.246 [.208,.284] & 1.00 & \(-44.8\) \\
Fixed repair \(K{=}3\) & 0.122 [.094,.152] & 3.00 & \(-57.2\) \\
Fixed repair \(K{=}5\) & 0.116 [.088,.144] & 5.00 & \(-57.8\) \\
Accepted-first & 0.690 [.648,.730] & 0.92 & \(-0.4\) \\
Last-accepted & 0.504 [.460,.546] & 2.28 & \(-19.0\) \\
Verifier-best-of-trajectory & 0.696 [.654,.736] & 0.73 & \(+0.2\) \\
VRR-Stop & \textbf{0.722} [.682,.760] & 0.72 & \(+2.8\) \\
\midrule
TPM reference & 0.694 [.652,.734] & 0.89 & 0.0 \\
\bottomrule
\end{tabular}
\end{table*}

\section{Full Calibration and Identifiability Results}
\label{app:full-identifiability}

This appendix examines when the parameter estimates can be trusted, when they fail, and locates the mechanism of failure.

\begin{figure}[t]
\centering
\includegraphics[width=\columnwidth]{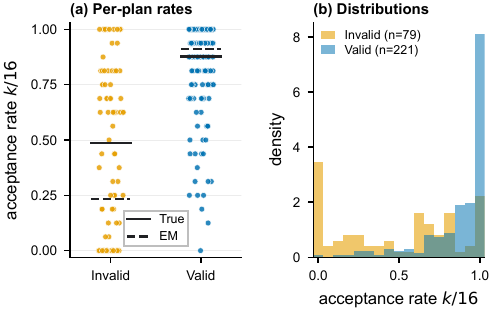}
\caption{Per-plan judge acceptance on the \(N{=}300\) probe window (\(M{=}16\)), split by true validity, shown as per-plan rates with class means in panel (a) and as class distributions in panel (b). Solid lines mark the true mean acceptance of each class and dashed lines the label-free EM estimates; EM recovers the valid-class rate almost exactly (\(0.91\) vs \(0.88\)) but underestimates the invalid-class rate (\(0.23\) vs \(0.49\)), the dominant source of downstream calibration error. The two classes overlap yet remain separable, which is why weakly supervised calibration is feasible but fragile.}
\label{fig:probe-separation}
\end{figure}

\textbf{Cross-family calibration stress.} Table~\ref{tab:app-calibration-stress} shows that the dominant calibration bias points in a different direction for each of the four verifiers: the Qwen family underestimates the false-acceptance rate, Llama underestimates the false-rejection rate, and the false-acceptance bias of Mistral reaches \(0.66\). Decision consequences are decoupled from bias magnitude: the largest bias, on Mistral, costs only \(0.3\) percentage points, whereas the smaller bias on Llama produces a \(58.0\)-point collapse; the dividing line is the discrimination ability \(J\), not the bias itself. Fig.~\ref{fig:probe-separation} supplies the upstream evidence for all of this: on the probe window, repeated verification yields acceptance-rate distributions for the two classes that overlap yet remain separable, which is why weakly supervised calibration is feasible---and why it is fragile.

\textbf{Identifiability collapse on Llama.} Table~\ref{tab:app-llama-em} exposes the failure mechanism. Binomial expectation-maximization (EM) yields \(\hat\rho_1=0.27\) at \(N=120\), yet increasing the sample to \(300\) worsens the estimate to \(0.077\): as \(J\) approaches zero the likelihood surface flattens, and additional samples only make EM converge more confidently to a degenerate solution. The capped beta-binomial estimator gives \(0.620\) at \(N=120\) but is unstable across random seeds. A parameter sweep under the deployed rule completes the picture: calibrated decisions recover the stop-early optimum only when \(\hat\rho_1\) is no lower than \(0.30\) to \(0.35\), with the midpoint crossing at \(\hat\rho_1\approx0.17\), so the EM estimate of \(0.077\) sits deep in the collapse zone.

\begin{table*}[t]
\centering
\small
\caption{Cross-family calibration stress (per-row \(N\) is each trace's
evaluation window, \(J\) uses the family-level convention, and Gap is
\(V_{\text{stop}}-V_{\text{TPM}}\) in percentage points; the TPM reference
is not a hindsight upper bound). Dominant bias reports the largest EM
estimation error as estimate minus truth, so negative values are
underestimates. Decision damage tracks \(J\) rather than the size of the
dominant calibration bias.}
\label{tab:app-calibration-stress}
\begin{tabular}{llcclccc}
\toprule
Judge & Family & \(N\) & \(J\) & Dominant bias & \(V_{\text{stop}}\) & \(V_{\text{TPM}}\) & Gap (pp) \\
\midrule
Qwen2.5-7B & Qwen & 200 & 0.23 & \(\Delta\rho_0=-0.29\) & 0.875 & 0.875 & \(+0.0\) \\
Qwen2.5-3B & Qwen & 500 & 0.46 & \(\Delta\rho_0=-0.19\) & 0.722 & 0.694 & \(+2.8\) \\
Mistral-7B & Mistral & 300 & 0.18 & \(\Delta\rho_0=-0.66\) & 0.463 & 0.467 & \(-0.3\) \\
Llama-3-8B & Llama & 300 & \textbf{0.03} & \(\Delta\rho_1=-0.53\) & \textbf{0.223} & \textbf{0.803} & \(\mathbf{-58.0}\) \\
\bottomrule
\end{tabular}
\end{table*}

\begin{table}[h]
\centering
\small
\caption{Identifiability collapse on the Llama-3-8B judge (target
\(\rho_1=0.609\)). Under the deployed rule, a parameter sweep shows
calibrated stopping recovers the stop-early optimum only for
\(\hat\rho_1\gtrsim0.30\)--\(0.35\) with the midpoint crossing at
\(\hat\rho_1\approx0.17\), so the EM estimate of \(0.077\) sits deep in
the collapse zone.}
\label{tab:app-llama-em}
\begin{tabular}{lcc}
\toprule
Estimator & \(\hat\rho_1\) (\(N{=}120\)) & \(\hat\rho_1\) (\(N{=}300\)) \\
\midrule
Binomial EM & 0.27 & 0.077 \\
Capped beta-binomial EM & 0.620 & unstable (seed) \\
\bottomrule
\end{tabular}
\end{table}

Stopping-decision agreement numbers for the estimator-input ablation are given in Appendix~\ref{app:ablations}; the flip-rate and identifiable-fraction grids underlying Fig.~\ref{fig:identifiability} are released with the anonymous code repository.

\section{Full Robustness Results}
\label{app:full-robustness}

This appendix examines whether the gains of VRR-Guard are confined to a single scenario and delineates the boundaries of its costs.

\begin{figure}[t]
\centering
\includegraphics[width=\columnwidth]{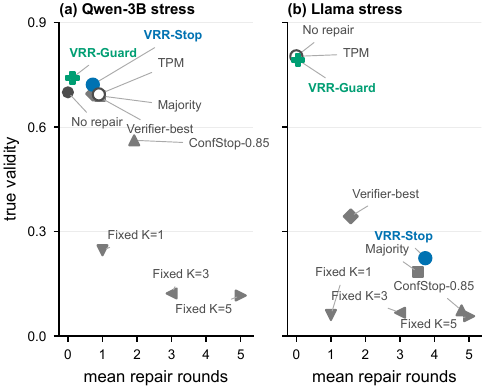}
\caption{True validity versus mean repair rounds for each stopping strategy under a working calibration (a, Qwen-3B stress) and a failed one (b, Llama stress); up-left is better, and the hollow marker is the ground-truth-parameter diagnostic TPM rather than a deployable strategy. Fixed-budget repair forms a dominated slide that pays more rounds for less validity, and in panel (b) calibrated VRR-Stop itself falls into that region while VRR-Guard stays at the no-repair corner. The pair thus covers both regimes, with calibrated stopping earning the low-cost, high-validity corner when calibration holds and the guarded fallback keeping that corner when it fails.}
\label{fig:pareto}
\end{figure}

\textbf{Full results.} Table~\ref{tab:app-full-robustness} extends Table~\ref{tab:robustness-main} with the TPM reference column and replay bootstrap intervals for Guard. VRR-Guard stays near the no-repair level in all seven settings, whereas fixed \(5\)-round repair degrades sharply in every setting except favorable. The Mistral row also provides a direct counterexample showing that the TPM reference is not a hindsight upper bound: its \(0.467\) falls below the no-repair \(0.507\). Fig.~\ref{fig:pareto} places the repair cost and true validity of two representative scenarios on the same plane.

\begin{table*}[t]
\centering
\small
\caption{Full guarded-stopping results with the TPM reference column and
replay bootstrap CIs; shift codes M, V, and T mark a change of model
family, verifier family, or task relative to the Qwen-3B baseline. The
TPM reference is a diagnostic rather than a hindsight upper bound (note
Mistral: TPM 0.467 \(<\) No repair 0.507). VRR-Guard stays within a few
points of no-repair in every row, including the Llama row where calibrated
stopping collapses.}
\label{tab:app-full-robustness}
\begin{tabular}{lccccccc}
\toprule
Setting & \(J\) & No repair & Fixed \(K{=}5\) & VRR-Stop & TPM ref. & VRR-Guard & Guard 95\% CI \\
\midrule
Qwen-3B favorable (--) & 0.39 & 0.740 & 0.875 & 0.845 & 0.850 & 0.810 & [.755,.865] \\
Qwen-3B stress (--)    & 0.46 & 0.700 & 0.116 & 0.722 & 0.694 & 0.742 & [.702,.780] \\
Qwen-7B stress (M)     & 0.23 & 0.875 & 0.075 & 0.875 & 0.875 & 0.875 & [.825,.920] \\
Mistral-7B stress (V)  & 0.18 & 0.507 & 0.047 & 0.463 & 0.467 & 0.487 & [.430,.543] \\
Llama-3-8B stress (M)  & 0.03 & 0.803 & 0.057 & 0.223 & 0.803 & 0.793 & [.747,.837] \\
MATH-500 (T)           & 0.22 & 0.798 & 0.150 & 0.798 & 0.798 & 0.796 & [.760,.832] \\
BFCL single (T)        & 0.07 & 0.812 & 0.372 & 0.782 & 0.780 & 0.810 & [.770,.848] \\
\bottomrule
\end{tabular}
\end{table*}

\textbf{Honest boundaries.} Guard does not dominate everywhere. On Mistral and BFCL it falls below no-repair by \(2.0\) and \(0.3\) percentage points, with paired intervals of \([-3.7,-0.7]\) and \([-1.3,+0.5]\) respectively, the latter containing zero (Appendix~\ref{app:paired-tests}). Under the favorable setting, Guard's \(0.81\) is below the \(0.875\) of fixed \(5\)-round repair---the opportunity cost of conservatism in a scenario where repair helps.

\textbf{Role of the retention margin.} Compared with margin-free Verifier-best on the same trajectories, Guard is \(45.0\) percentage points higher in the Llama scenario (95\% CI \([+39.0,+51.0]\)), while the differences in the remaining settings lie between \(-4.0\) and \(+8.0\) percentage points. The retention margin \(\delta\) is nearly neutral in most scenarios; its value is realized precisely where the verifier signal is least trustworthy.

\textbf{A no-collapse counterexample, and cost.} Under the near-inert repair operator of BFCL multi-turn (\(\alpha=0.02\), \(\beta=0.04\)), all strategies coincide within confidence intervals: multi-round repair is not inherently harmful, and severe degradation appears only together with a high break probability. Regarding cost, each round consumes a fixed \(M\) verification calls and at most one repair call, so the cost ordering of the strategies is exactly the \(\langle K\rangle\) ordering reported in each table; VRR-Guard operates by replaying saved trajectories and incurs no additional verification calls.

\section{Paired Statistical Tests}
\label{app:paired-tests}

This appendix provides same-trajectory paired statistical support for the percentage-point claims in the main text. All comparisons are paired per instance on the same batch of task instances and saved repair trajectories; differences in true validity are assessed with \(10{,}000\) instance-level bootstrap resamples, directional binary comparisons use exact McNemar tests, and a positive \(\Delta V\) indicates that the left-hand strategy is better on the same instances. The decision sequences of VRR-Stop and the TPM reference follow the same cross-fitting protocol as the main text.

\begin{table*}[t]
\centering
\small
\setlength{\tabcolsep}{4pt}
\caption{Paired replay statistics for load-bearing comparisons.
Positive \(\Delta V\) means the left strategy has higher true validity
on the same saved instances. Every headline percentage claim in the main
text maps to one row of this table.}
\label{tab:paired-stats}
\begin{tabular}{llcc}
\toprule
Section / setting & Comparison & \(\Delta V\) [95\% paired CI] & Paired test \\
\midrule
RQ1 interior peak & round 2 vs final round & \(+0.747\,[+0.693,+0.797]\) & paired bootstrap \\
RQ1 interior peak & round 0 vs final round & \(+0.327\,[+0.257,+0.397]\) & paired bootstrap \\
RQ2 Qwen-3B stress & VRR-Stop vs Fixed-\(K{=}5\) & \(+0.606\,[+0.560,+0.650]\) & McNemar \(p<2\times10^{-85}\) \\
RQ2 Qwen-3B stress & VRR-Stop vs ConfStop-0.85 & \(+0.160\,[+0.126,+0.194]\) & McNemar \(p<1\times10^{-18}\) \\
RQ2 Qwen-3B stress & VRR-Stop vs Majority stopping & \(+0.032\,[+0.016,+0.050]\) & McNemar \(p<2\times10^{-4}\) \\
RQ2 Qwen-3B stress & VRR-Stop vs TPM reference & \(+0.028\,[+0.014,+0.044]\) & McNemar \(p<6\times10^{-4}\) \\
RQ2 Qwen-3B stress & VRR-Stop vs No repair & \(+0.022\,[-0.004,+0.048]\) & paired bootstrap \\
RQ3 Llama stress & VRR-Stop vs TPM reference & \(-0.580\,[-0.637,-0.520]\) & McNemar \(p<2\times10^{-48}\) \\
RQ4 Llama stress & VRR-Guard vs failed VRR-Stop & \(+0.570\,[+0.510,+0.627]\) & McNemar \(p<1\times10^{-47}\) \\
RQ4 Llama stress & VRR-Guard vs Fixed-\(K{=}5\) & \(+0.737\,[+0.683,+0.787]\) & McNemar \(p<2\times10^{-62}\) \\
RQ4 Llama stress & VRR-Guard vs Verifier-best (unguarded) & \(+0.450\,[+0.390,+0.510]\) & paired bootstrap \\
RQ4 Mistral stress & VRR-Guard vs No repair & \(-0.020\,[-0.037,-0.007]\) & paired bootstrap \\
RQ4 BFCL & VRR-Guard vs No repair & \(-0.003\,[-0.013,+0.005]\) & paired bootstrap \\
\bottomrule
\end{tabular}
\end{table*}

\section{Extended Discussion}
\label{app:extended-discussion}

This appendix elaborates on applicability conditions and method boundaries that the main text does not cover in detail, including how harmful repair arises, how calibration error affects stopping decisions, the failure modes of different verifiers, and the benefits and costs of the conservative fallback strategy. A common thread across the subsections is that calibration quality should be measured by whether the decision sign flips, not by the absolute error of parameter estimates.

\subsection{Repair Benefit and Damage Risk}

Whether to continue repairing depends on the relative magnitude of the repair benefit and the damage loss, not on the number of repair rounds already executed. By Eq.~\eqref{eq:gated-gain}, the true marginal gain of a single repair round is
\[
G_k=(1-b_k)\alpha-b_k\beta.
\]
The first term captures the expected benefit of an invalid plan becoming valid through repair, and the second term captures the expected loss of a valid plan being damaged by repair. Even under the same maximum repair budget, different values of \(b_k,\alpha,\beta\) yield different optimal stopping times; verifier noise \(\rho_0,\rho_1\) shifts this timing indirectly through the posterior belief \(b_k\).

The results in Appendix~\ref{app:full-dynamics} exhibit pronounced differences across settings. In the GSM8K stress setting and on MATH-500, the repairer rarely corrects invalid plans yet frequently damages valid ones: fixed five-round repair reduces true validity from \(0.700\) to \(0.116\) and from \(0.798\) to \(0.150\), respectively. BFCL multi-turn instead shows nearly inert repair dynamics, with \(\alpha=0.02\) and \(\beta=0.04\), and no statistically distinguishable difference among stopping policies. This counterexample shows that multi-round repair is not inherently harmful; substantial degradation emerges only when the repairer has a non-negligible capacity to damage valid states.

Verifier discrimination ability and repairer safety also require separate examination. On GSM8K, the process reward model verifier Qwen2.5-Math-PRM-7B~\citep{zhang2025lessons} attains a Youden's index of \(J=0.805\) (this setting uses Qwen2.5-3B as the generator and is independent of the MATH-500 row in Table~\ref{tab:loop-dynamics}), yet the stronger verifier does not prevent the repairer from damaging correct plans. In this setting, fixed five-round repair still reduces validity from \(0.727\) to \(0.097\), whereas the true-parameter myopic (TPM) reference remains at \(0.663\). Improving verifier quality therefore strengthens the reliability of stopping decisions but cannot eliminate the risk introduced by the repair operation itself.

\subsection{Non-Stationarity and Interior Peaks}

The repairer's state-transition probabilities may change across rounds. In the favorable setting of Qwen2.5-3B, the repair success rate under the per-round convention drops from \(0.415\) in the first round to \(0.032\) in the fifth (Table~\ref{tab:loop-dynamics} reports a first-round value of \(0.423\) under the per-call convention), while the damage probability stays low throughout. This pattern admits a natural interpretation: easily repairable errors are eliminated first, and the remaining errors are harder to fix. Describing the entire repair process with transition probabilities estimated from the first round should therefore be regarded only as a locally stationary approximation.

The non-stationarity diagnostic trajectory in the main text's loop-dynamics experiment further illustrates the interior peak that repair can produce. Early repair completes initial solutions truncated by the generation-length limit, raising true validity to \(0.87\) at round two. After context drift is introduced at round three, true validity gradually declines to \(0.12\). This trajectory demonstrates that the optimal submission time can occur at an intermediate round, but it does not imply that a stopping policy based on stationary parameters can necessarily locate that peak. In paired replay, VRR-Stop and the TPM reference attain \(0.693\) and \(0.720\) (cross-fitted convention), both far below the \(0.867\) obtained by post hoc selection of round two.

This result delineates the capability boundary of the current method. VRR-Stop judges whether one more repair round still yields a positive gain under the current local model; it does not search the full trajectory for the post hoc optimal round. When the repair mechanism changes abruptly at some round, \(\alpha\) and \(\beta\) estimated from the first round cannot represent this change in advance. The interior-peak experiment thus both confirms the necessity of the stopping problem and indicates room for further research on round-dependent dynamic models.

\subsection{Decision Robustness under Calibration Error}

Parameter estimation error affects the stopping action only when it changes the sign of \(G_k-\tau\). For Mistral-7B, the estimation bias of the false-acceptance rate reaches \(\Delta\rho_0=-0.66\), yet VRR-Stop attains a true validity of \(0.463\), within \(0.3\) percentage points of the TPM reference at \(0.467\). Despite the large parameter error, the corresponding marginal gain stays in the same decision region, so the final action does not change.

Llama-3-8B exhibits the opposite outcome. Its verifier discrimination ability is only \(J=0.03\), and the binomial-mixture estimate underestimates the false-rejection rate by about \(0.53\), which systematically inflates the posterior belief \(b_k\) and thus substantially underestimates the damage risk represented by \(b_k\beta\). The estimation error eventually crosses the stopping boundary, driving VRR-Stop down to \(0.223\) from the TPM reference of \(0.803\). The direct cause of the degradation is not the parameter bias itself but the fact that the bias flips the decision sign between continuing and stopping.

BFCL single-call provides further corroboration. Its verifier has a \(J\) of only \(0.07\), but the states it visits carry large decision margins, and VRR-Stop differs from the TPM reference by only \(0.3\) percentage points. A uniform trust threshold therefore cannot be set from the \(J\) value alone. The controlled experiment shown in Fig.~\ref{fig:identifiability} likewise indicates that when low \(J\) coincides with a low decision margin \(\Delta\), the sign-flip rate reaches \(0.183\), whereas it is only \(0.014\) in regions with high \(J\) or high \(\Delta\). The reliability of stopping decisions is jointly determined by verifier discrimination ability, decision margin, and calibration sample size.

\subsection{Family-Conditional Verifier Failures}

The current experiments further show that different model families can exhibit different directions of verifier failure. Across the eight Qwen configurations, the binomial-mixture estimate underestimates the false-acceptance rate in every case, with \(\Delta\rho_0\) ranging from \(-0.16\) to \(-0.31\) (the full list of biases is released with the anonymous code repository). For Llama-3-8B, the false-acceptance rate is estimated essentially accurately, but the false-rejection rate is markedly underestimated (by about \(0.53\) at \(N{=}300\)). Mistral-7B instead shows a pronounced tendency toward false acceptance, with a true false-acceptance rate of about \(0.71\), yet its large decision margin prevents the stopping sign from flipping.

These results indicate that verifier noise should not be treated as a fixed bias transferable across models. The same calibration method can underestimate the false-acceptance rate in one model family and the false-rejection rate in another, and the two biases propagate through the posterior belief into Eq.~\eqref{eq:gated-gain} in different directions. That said, the existing experiments cover only three model families, and some configurations share tasks and samples. These findings should therefore be read as family-conditional patterns observed within the current experimental scope, not as universal laws determined by model architecture.

The process reward model and the executors used in our experiments also impose different calibration conditions. A deterministic verifier produces no new stochastic observations under repeated queries on the same plan, so the binomial-mixture estimate based on repeated acceptance counts cannot be applied directly. The stopping criterion itself remains applicable in this case, but \(\rho_0\) and \(\rho_1\) must be estimated directly from a small number of labeled samples.

\subsection{Safety and Upside of the Guarded Fallback}

When Eq.~\eqref{eq:sign-identifiability} fails to hold, VRR-Guard no longer relies on precise noise parameters; instead, it raises the evidential strength required to replace the current candidate. The policy retains the best candidate so far and performs a replacement only when the verification acceptance count of a new candidate exceeds that of the current one by at least \(\delta\). It does not prevent the system from generating subsequent candidates; it avoids committing a possibly degraded repair result when the evidence is insufficient.

In the Llama-3-8B scenario, VRR-Guard restores true validity to \(0.793\), an improvement of \(57.0\) percentage points over the failed VRR-Stop and \(73.7\) percentage points over fixed five-round repair. When unsupervised calibration approaches non-identifiability, retaining the existing candidate is thus more robust than continuing to rely on erroneous parameters.

This robustness comes with an observable opportunity cost. In the favorable setting, VRR-Guard reaches a validity of \(0.810\), below the \(0.875\) of fixed five-round repair. In the Mistral-7B and BFCL settings, VRR-Guard falls short of no-repair by \(2.0\) and \(0.3\) percentage points, respectively, and the paired interval for BFCL contains zero. VRR-Guard therefore carries no theoretical guarantee of strictly dominating no-repair. It is better viewed as a conservative fallback close to the no-repair baseline: it bounds the downside risk when calibration is unreliable, at the cost of forgoing part of the potential benefit in scenarios where repair is beneficial.

\subsection{Scope and Deployment Implications}

This work represents whether a plan is truly valid with a binary state. This binary validity representation suits tasks with well-defined labels, such as answer correctness, tool executability, and constraint satisfaction, but it cannot fully describe partial correctness, multiple error types, or long-horizon plans with staged goals. For such tasks, the state space needs to be extended to multi-level or structured representations, and the corresponding stopping boundary may no longer be given by a single threshold.

The strict condition \(\alpha+\beta\le1\) holds on seven of the eight empirical trajectories, but the GSM8K stress trajectory of Qwen2.5-3B clearly violates it. Conclusions on that trajectory therefore rest on single-round marginal gains and finite-trajectory diagnostics rather than on global monotonicity. The TPM reference likewise serves only to isolate decision differences caused by parameter calibration; it is not a globally optimal upper bound over the finite horizon. Post hoc selection of the best round can be used to analyze trajectory structure but cannot serve as a deployable policy.

Deploying VRR-Stop should begin with estimating the noise parameters and their error ranges from repeated verification records and a small number of labeled transition samples. Only when \(\hat G_k-B_k>\tau\) or \(\hat G_k+B_k\le\tau\) does the available evidence suffice to support a definite continue or stop action. If the confidence interval straddles the stopping boundary, or if the model, verifier, or task distribution shifts, the system should switch to VRR-Guard or recollect labeled samples for recalibration. Existing parameters should not be transferred directly to a new model family, verifier, or repair prompt.

Future work can model round-dependent \(\alpha_t\) and \(\beta_t\) to detect late context drift and interior peaks. The binary true state can also be extended to multi-level plan quality, describing partial correctness and distinct error types. Another direction is to estimate instance-conditional noise that depends on plan features, so that plans of different difficulty receive different stopping boundaries. Finally, per-instance marginal gains can support cross-task compute allocation, selecting the instances most worth further repair when the total inference budget is limited.
\section{Additional Ablations}
\label{app:ablations}

This appendix reports sensitivity analyses that have already been run and can be replayed at zero cost.

\textbf{Calibration-input ablation.} In the GSM8K stress setting, the final validity under three calibration inputs---binomial expectation-maximization (EM), method of moments (MoM), and labeled maximum-likelihood estimation (MLE)---is \(0.722\), \(0.690\), and \(0.690\), respectively, with the differences in stopping decisions concentrated on a few borderline instances. Beta-binomial EM (BB-EM) yields \(0.666\) (95\% CI \([0.624,0.706]\), \(\langle K\rangle=1.26\)): its overly flexible fit pushes \(\hat\rho_0\) up to \(0.508\), which systematically depresses the posterior belief and induces excessive repair. Table~\ref{tab:app-llama-em} provides a comparative diagnosis of the two EM variants under within-class heterogeneity.

\textbf{Retention-margin ablation.} A pure replay sweep over \(\delta\in\{0,\ldots,8\}\) (\(M=8\)) across all seven settings shows that \(\delta=5\) is the most robust choice, trailing the per-setting optimal \(\delta\) by at most \(4.0\) percentage points in any setting.

\textbf{Calibration-sample-size sensitivity.} Under subsampled re-estimation with \(N_{\text{calib}}\in\{50,100,200,300\}\), the decision agreement rate rises from \(0.973\) to \(0.997\), the sign-flip rate falls from \(0.016\) to \(0.002\), and the absolute validity gap to the TPM reference shrinks from \(0.012\) to \(0.001\). Systematic ablations of the verification budget \(M\) and the gain threshold \(\tau\) were not run within the scope of this work and are left for future work.

\section{Reproducibility Details}
\label{app:reproducibility}

All experiments were run on a single server equipped with eight NVIDIA A800 80GB PCIe GPUs, with each experiment completed on a single GPU. The software environment comprises Ubuntu 22.04, CUDA 12.8, PyTorch 2.10.0 (cuDNN 9.10.2), and the inference engine vLLM 0.19.1~\citep{kwon2023pagedattention}. The temperature for generation, judging, and repair is \(0.7\), with the repair temperature raised to \(1.0\) in the stress setting. Code versions, random seeds, the complete generation, judging, and repair prompts for each task, and an index of raw data files are released with the anonymous code repository. True validity is adjudicated by exact match on GSM8K, symbolic verification on MATH-500, sandboxed unit tests on MBPP (\(5\)-second timeout), and abstract syntax tree (AST) matching plus the official executor on BFCL; Appendix~\ref{app:full-setup} describes how the generator prior is obtained.

Per-instance LLM trajectories are single stochastic realizations and are not bit-wise reproducible across inference-engine versions. All numbers in this paper are computed by deterministic replay of frozen trajectories and do not depend on regeneration.

\bibliography{vrr}

\begin{thebibliography}{38}
\providecommand{\natexlab}[1]{#1}

\bibitem[{Austin et~al.(2021)Austin, Odena, Nye, Bosma, Michalewski, Dohan,
  Jiang, Cai, Terry, Le, and Sutton}]{austin2021program}
Austin, J.; Odena, A.; Nye, M.; Bosma, M.; Michalewski, H.; Dohan, D.; Jiang,
  E.; Cai, C.; Terry, M.; Le, Q.; and Sutton, C. 2021.
\newblock Program Synthesis with Large Language Models.
\newblock arXiv:2108.07732.

\bibitem[{Cobbe et~al.(2021)Cobbe, Kosaraju, Bavarian, Chen, Jun, Kaiser,
  Plappert, Tworek, Hilton, Nakano, Hesse, and Schulman}]{cobbe2021training}
Cobbe, K.; Kosaraju, V.; Bavarian, M.; Chen, M.; Jun, H.; Kaiser, L.; Plappert,
  M.; Tworek, J.; Hilton, J.; Nakano, R.; Hesse, C.; and Schulman, J. 2021.
\newblock Training Verifiers to Solve Math Word Problems.
\newblock arXiv:2110.14168.

\bibitem[{Collot et~al.(2026)Collot, Fraser, Zhao, Shen, Willi, and
  Leontiadis}]{collot2025balanced}
Collot, S.; Fraser, C.; Zhao, J.; Shen, W.~F.; Willi, T.; and Leontiadis, I.
  2026.
\newblock Balanced Accuracy: The Right Metric for Evaluating {LLM} Judges -
  Explained through Youden{'}s {J} statistic.
\newblock In Matusevych, Y.; Eryi{\u{g}}it, G.; and Aletras, N., eds.,
  \emph{Proceedings of the 19th Conference of the {E}uropean Chapter of the
  {A}ssociation for {C}omputational {L}inguistics (Volume 5: Industry Track)},
  927--936. Rabat, Morocco: Association for Computational Linguistics.
\newblock ISBN 979-8-89176-384-5.

\bibitem[{Dorner et~al.(2026)Dorner, Chen, Cruz, and
  Yang}]{dorner2025rocnreroll}
Dorner, F.~E.; Chen, Y.; Cruz, A.~F.; and Yang, F. 2026.
\newblock {ROC}-n-reroll: How Verifier Imperfection Affects Test-Time Scaling.
\newblock In \emph{The Fourteenth International Conference on Learning
  Representations (ICLR)}.

\bibitem[{Feng et~al.(2026)Feng, Shen, Balashankar, Gerner-Beuerle, and
  Rodrigues}]{feng2026noisy}
Feng, C.; Shen, M.; Balashankar, A.; Gerner-Beuerle, C.; and Rodrigues, M.
  R.~D. 2026.
\newblock Noisy but Valid: Robust Statistical Evaluation of {LLM}s with
  Imperfect Judges.
\newblock In \emph{The Fourteenth International Conference on Learning
  Representations (ICLR)}.
\newblock ArXiv:2601.20913.

\bibitem[{Fu et~al.(2026)Fu, Wang, Tian, and Zhao}]{fu2026deepconf}
Fu, Y.; Wang, X.; Tian, Y.; and Zhao, J. 2026.
\newblock Deep Think with Confidence.
\newblock In \emph{The Fourteenth International Conference on Learning
  Representations (ICLR)}.

\bibitem[{Gao, Schulman, and Hilton(2023)}]{gao2023scaling}
Gao, L.; Schulman, J.; and Hilton, J. 2023.
\newblock Scaling Laws for Reward Model Overoptimization.
\newblock In \emph{Proceedings of the 40th International Conference on Machine
  Learning}, volume 202 of \emph{Proceedings of Machine Learning Research}.

\bibitem[{Gou et~al.(2024)Gou, Shao, Gong, Shen, Yang, Duan, and
  Chen}]{gou2024critic}
Gou, Z.; Shao, Z.; Gong, Y.; Shen, Y.; Yang, Y.; Duan, N.; and Chen, W. 2024.
\newblock {CRITIC}: Large Language Models Can Self-Correct with
  Tool-Interactive Critiquing.
\newblock In \emph{The Twelfth International Conference on Learning
  Representations (ICLR)}. OpenReview.net.

\bibitem[{Grattafiori et~al.(2024)Grattafiori, Dubey, Jauhri, Pandey, Kadian,
  Al-Dahle et~al.}]{grattafiori2024llama3}
Grattafiori, A.; Dubey, A.; Jauhri, A.; Pandey, A.; Kadian, A.; Al-Dahle, A.;
  et~al. 2024.
\newblock The {Llama} 3 Herd of Models.
\newblock \emph{CoRR}, abs/2407.21783.

\bibitem[{Hendrycks et~al.(2021)Hendrycks, Burns, Kadavath, Arora, Basart,
  Tang, Song, and Steinhardt}]{hendrycks2021measuring}
Hendrycks, D.; Burns, C.; Kadavath, S.; Arora, A.; Basart, S.; Tang, E.; Song,
  D.; and Steinhardt, J. 2021.
\newblock Measuring Mathematical Problem Solving With the {MATH} Dataset.
\newblock In Vanschoren, J.; and Yeung, S., eds., \emph{Proceedings of the
  Neural Information Processing Systems Track on Datasets and Benchmarks 1
  (NeurIPS Datasets and Benchmarks 2021)}.

\bibitem[{Huang et~al.(2024)Huang, Chen, Mishra, Zheng, Yu, Song, and
  Zhou}]{huang2023large}
Huang, J.; Chen, X.; Mishra, S.; Zheng, H.~S.; Yu, A.~W.; Song, X.; and Zhou,
  D. 2024.
\newblock Large Language Models Cannot Self-Correct Reasoning Yet.
\newblock In \emph{The Twelfth International Conference on Learning
  Representations (ICLR)}. OpenReview.net.

\bibitem[{Jiang et~al.(2023)Jiang, Sablayrolles, Mensch, Bamford, Chaplot,
  de~las Casas, Bressand, Lengyel, Lample, Saulnier, Lavaud, Lachaux, Stock,
  Le~Scao, Lavril, Wang, Lacroix, and El~Sayed}]{jiang2023mistral7b}
Jiang, A.~Q.; Sablayrolles, A.; Mensch, A.; Bamford, C.; Chaplot, D.~S.; de~las
  Casas, D.; Bressand, F.; Lengyel, G.; Lample, G.; Saulnier, L.; Lavaud,
  L.~R.; Lachaux, M.-A.; Stock, P.; Le~Scao, T.; Lavril, T.; Wang, T.; Lacroix,
  T.; and El~Sayed, W. 2023.
\newblock Mistral 7B.
\newblock \emph{CoRR}, abs/2310.06825.

\bibitem[{Kamoi et~al.(2024{\natexlab{a}})Kamoi, Das, Lou, Ahn, Zhao, Lu,
  Zhang, Zhang, Zhang, Vummanthala, Dave, Qin, Cohan, Yin, and
  Zhang}]{kamoi2024evaluating}
Kamoi, R.; Das, S. S.~S.; Lou, R.; Ahn, J.~J.; Zhao, Y.; Lu, X.; Zhang, N.;
  Zhang, Y.; Zhang, R.~H.; Vummanthala, S.~R.; Dave, S.; Qin, S.; Cohan, A.;
  Yin, W.; and Zhang, R. 2024{\natexlab{a}}.
\newblock Evaluating {LLM}s at Detecting Errors in {LLM} Responses.
\newblock In \emph{First Conference on Language Modeling (COLM)}.

\bibitem[{Kamoi et~al.(2024{\natexlab{b}})Kamoi, Zhang, Zhang, Han, and
  Zhang}]{kamoi2024when}
Kamoi, R.; Zhang, Y.; Zhang, N.; Han, J.; and Zhang, R. 2024{\natexlab{b}}.
\newblock When Can {LLM}s Actually Correct Their Own Mistakes? {A} Critical
  Survey of Self-Correction of {LLM}s.
\newblock \emph{Transactions of the Association for Computational Linguistics},
  12: 1417--1440.

\bibitem[{Khalaf et~al.(2025)Khalaf, Verdun, Oesterling, Lakkaraju, and
  Calmon}]{khalaf2025inference}
Khalaf, H.; Verdun, C.~M.; Oesterling, A.; Lakkaraju, H.; and Calmon, F. d.~P.
  2025.
\newblock Inference-Time Reward Hacking in Large Language Models.
\newblock In \emph{Advances in Neural Information Processing Systems 38
  (NeurIPS 2025)}.
\newblock Spotlight.

\bibitem[{Khalifa et~al.(2026)Khalifa, Agarwal, Logeswaran, Kim, Peng, Lee,
  Lee, and Wang}]{khalifa2025process}
Khalifa, M.; Agarwal, R.; Logeswaran, L.; Kim, J.; Peng, H.; Lee, M.; Lee, H.;
  and Wang, L. 2026.
\newblock Process Reward Models That Think.
\newblock \emph{Transactions on Machine Learning Research}.

\bibitem[{Kumar et~al.(2025)Kumar, Zhuang, Agarwal, Su, Co-Reyes, Singh,
  Baumli, Iqbal, Bishop, Roelofs, Zhang, McKinney, Shrivastava, Paduraru,
  Tucker, Precup, Behbahani, and Faust}]{kumar2024training}
Kumar, A.; Zhuang, V.; Agarwal, R.; Su, Y.; Co-Reyes, J.~D.; Singh, A.; Baumli,
  K.; Iqbal, S.; Bishop, C.; Roelofs, R.; Zhang, L.~M.; McKinney, K.;
  Shrivastava, D.; Paduraru, C.; Tucker, G.; Precup, D.; Behbahani, F.; and
  Faust, A. 2025.
\newblock Training Language Models to Self-Correct via Reinforcement Learning.
\newblock In \emph{The Thirteenth International Conference on Learning
  Representations (ICLR)}. OpenReview.net.

\bibitem[{Kwon et~al.(2023)Kwon, Li, Zhuang, Sheng, Zheng, Yu, Gonzalez, Zhang,
  and Stoica}]{kwon2023pagedattention}
Kwon, W.; Li, Z.; Zhuang, S.; Sheng, Y.; Zheng, L.; Yu, C.~H.; Gonzalez, J.~E.;
  Zhang, H.; and Stoica, I. 2023.
\newblock Efficient Memory Management for Large Language Model Serving with
  {PagedAttention}.
\newblock In Flinn, J.; Seltzer, M.~I.; Druschel, P.; Kaufmann, A.; and Mace,
  J., eds., \emph{Proceedings of the 29th Symposium on Operating Systems
  Principles (SOSP '23)}, 611--626. Koblenz, Germany: ACM.
\newblock ISBN 9798400702297.

\bibitem[{Lightman et~al.(2024)Lightman, Kosaraju, Burda, Edwards, Baker, Lee,
  Leike, Schulman, Sutskever, and Cobbe}]{lightman2023lets}
Lightman, H.; Kosaraju, V.; Burda, Y.; Edwards, H.; Baker, B.; Lee, T.; Leike,
  J.; Schulman, J.; Sutskever, I.; and Cobbe, K. 2024.
\newblock Let's Verify Step by Step.
\newblock In \emph{The Twelfth International Conference on Learning
  Representations (ICLR)}. OpenReview.net.

\bibitem[{Lu et~al.(2025)Lu, Teehan, Jin, and Ren}]{lu2025when}
Lu, J.; Teehan, R.; Jin, J.; and Ren, M. 2025.
\newblock When Does Verification Pay Off? {A} Closer Look at {LLM}s as Solution
  Verifiers.
\newblock arXiv:2512.02304.

\bibitem[{Madaan et~al.(2023)Madaan, Tandon, Gupta, Hallinan, Gao, Wiegreffe,
  Alon, Dziri, Prabhumoye, Yang, Gupta, Majumder, Hermann, Welleck,
  Yazdanbakhsh, and Clark}]{madaan2023selfrefine}
Madaan, A.; Tandon, N.; Gupta, P.; Hallinan, S.; Gao, L.; Wiegreffe, S.; Alon,
  U.; Dziri, N.; Prabhumoye, S.; Yang, Y.; Gupta, S.; Majumder, B.~P.; Hermann,
  K.; Welleck, S.; Yazdanbakhsh, A.; and Clark, P. 2023.
\newblock Self-Refine: Iterative Refinement with Self-Feedback.
\newblock In Oh, A.; Naumann, T.; Globerson, A.; Saenko, K.; Hardt, M.; and
  Levine, S., eds., \emph{Advances in Neural Information Processing Systems},
  volume~36, 46534--46594. Curran Associates, Inc.

\bibitem[{Pan et~al.(2024{\natexlab{a}})Pan, Jones, Jagadeesan, and
  Steinhardt}]{pan2024feedback}
Pan, A.; Jones, E.; Jagadeesan, M.; and Steinhardt, J. 2024{\natexlab{a}}.
\newblock Feedback Loops With Language Models Drive In-Context Reward Hacking.
\newblock In \emph{Proceedings of the 41st International Conference on Machine
  Learning}, volume 235 of \emph{Proceedings of Machine Learning Research},
  39154--39200. PMLR.

\bibitem[{Pan et~al.(2024{\natexlab{b}})Pan, He, Bowman, and
  Feng}]{pan2024spontaneous}
Pan, J.; He, H.; Bowman, S.~R.; and Feng, S. 2024{\natexlab{b}}.
\newblock Spontaneous Reward Hacking in Iterative Self-Refinement.
\newblock arXiv:2407.04549.

\bibitem[{Patil et~al.(2025)Patil, Mao, Cheng-Jie~Ji, Yan, Suresh, Stoica, and
  E.~Gonzalez}]{patil2025bfcl}
Patil, S.~G.; Mao, H.; Cheng-Jie~Ji, C.; Yan, F.; Suresh, V.; Stoica, I.; and
  E.~Gonzalez, J. 2025.
\newblock The Berkeley Function Calling Leaderboard ({BFCL}): From Tool Use to
  Agentic Evaluation of Large Language Models.
\newblock In \emph{Forty-second International Conference on Machine Learning}.

\bibitem[{Platanios, Blum, and Mitchell(2016)}]{platanios2016estimating}
Platanios, E.~A.; Blum, A.; and Mitchell, T.~M. 2016.
\newblock Estimating Accuracy from Unlabeled Data: A Bayesian Approach.
\newblock In \emph{Proceedings of the 33rd International Conference on Machine
  Learning}, volume~48 of \emph{Proceedings of Machine Learning Research}.

\bibitem[{Shinn et~al.(2023)Shinn, Cassano, Gopinath, Narasimhan, and
  Yao}]{shinn2023reflexion}
Shinn, N.; Cassano, F.; Gopinath, A.; Narasimhan, K.; and Yao, S. 2023.
\newblock Reflexion: Language Agents with Verbal Reinforcement Learning.
\newblock In Oh, A.; Naumann, T.; Globerson, A.; Saenko, K.; Hardt, M.; and
  Levine, S., eds., \emph{Advances in Neural Information Processing Systems},
  volume~36, 8634--8652. Curran Associates, Inc.

\bibitem[{Stroebl, Kapoor, and Narayanan(2026)}]{stroebl2024limits}
Stroebl, B.; Kapoor, S.; and Narayanan, A. 2026.
\newblock The Limits of Inference Scaling Through Resampling.
\newblock In \emph{The Fourteenth International Conference on Learning
  Representations (ICLR)}.

\bibitem[{Sun et~al.(2026)Sun, Cheng, Li, Chen, and Wang}]{sun2026stop}
Sun, R.; Cheng, W.; Li, D.; Chen, H.; and Wang, W. 2026.
\newblock Stop When Enough: Adaptive Early-Stopping for Chain-of-Thought
  Reasoning.
\newblock In \emph{Proceedings of the 64th Annual Meeting of the Association
  for Computational Linguistics}. San Diego, California, USA: Association for
  Computational Linguistics.
\newblock ArXiv:2510.10103.

\bibitem[{Wang et~al.(2023)Wang, Wei, Schuurmans, Le, Chi, Narang, Chowdhery,
  and Zhou}]{wang2022selfconsistency}
Wang, X.; Wei, J.; Schuurmans, D.; Le, Q.; Chi, E.; Narang, S.; Chowdhery, A.;
  and Zhou, D. 2023.
\newblock Self-Consistency Improves Chain of Thought Reasoning in Language
  Models.
\newblock In \emph{The Eleventh International Conference on Learning
  Representations (ICLR)}. OpenReview.net.

\bibitem[{Yang et~al.(2024)Yang, Yang, Zhang, Hui, Zheng, Yu, Li, Liu, Huang,
  Wei et~al.}]{qwen2024qwen25}
Yang, A.; Yang, B.; Zhang, B.; Hui, B.; Zheng, B.; Yu, B.; Li, C.; Liu, D.;
  Huang, F.; Wei, H.; et~al. 2024.
\newblock Qwen2.5 Technical Report.
\newblock arXiv:2412.15115.

\bibitem[{Yang et~al.(2025)Yang, Zhang, Wang, Xu, Lin, and
  Sui}]{yang2025probabilistic}
Yang, Z.; Zhang, Y.; Wang, Y.; Xu, Z.; Lin, J.; and Sui, Z. 2025.
\newblock A Probabilistic Inference Scaling Theory for {LLM} Self-Correction.
\newblock In \emph{Proceedings of the 2025 Conference on Empirical Methods in
  Natural Language Processing}, 13573--13587. Suzhou, China: Association for
  Computational Linguistics.

\bibitem[{Ye et~al.(2025)Ye, Melo, Kaddar, Blunsom, Staton, and
  Gal}]{ye2025uncertainty}
Ye, Z.; Melo, L.~C.; Kaddar, Y.; Blunsom, P.; Staton, S.; and Gal, Y. 2025.
\newblock Uncertainty-Aware Step-wise Verification with Generative Reward
  Models.
\newblock In \emph{ICLR 2025 Workshop on Quantify Uncertainty and Hallucination
  in Foundation Models}.

\bibitem[{Yu, Li, and Wang(2025)}]{yu2025scaling}
Yu, F.; Li, Y.; and Wang, B. 2025.
\newblock Scaling Flaws of Verifier-Guided Search in Mathematical Reasoning.
\newblock arXiv:2502.00271.

\bibitem[{Zhang et~al.(2025{\natexlab{a}})Zhang, Hosseini, Bansal, Kazemi,
  Kumar, and Agarwal}]{zhang2024generative}
Zhang, L.; Hosseini, A.; Bansal, H.; Kazemi, M.; Kumar, A.; and Agarwal, R.
  2025{\natexlab{a}}.
\newblock Generative Verifiers: Reward Modeling as Next-Token Prediction.
\newblock In \emph{The Thirteenth International Conference on Learning
  Representations (ICLR)}. OpenReview.net.

\bibitem[{Zhang et~al.(2025{\natexlab{b}})Zhang, Zheng, Wu, Zhang, Lin, Yu,
  Liu, Zhou, and Lin}]{zhang2025lessons}
Zhang, Z.; Zheng, C.; Wu, Y.; Zhang, B.; Lin, R.; Yu, B.; Liu, D.; Zhou, J.;
  and Lin, J. 2025{\natexlab{b}}.
\newblock The Lessons of Developing Process Reward Models in Mathematical
  Reasoning.
\newblock In Che, W.; Nabende, J.; Shutova, E.; and Pilehvar, M.~T., eds.,
  \emph{Findings of the Association for Computational Linguistics: ACL 2025},
  10495--10516. Vienna, Austria: Association for Computational Linguistics.
\newblock ISBN 979-8-89176-256-5.

\bibitem[{Zhao, Awasthi, and Gollapudi(2025)}]{zhao2025sample}
Zhao, E.; Awasthi, P.; and Gollapudi, S. 2025.
\newblock Sample, Scrutinize and Scale: Effective Inference-Time Search by
  Scaling Verification.
\newblock In \emph{Proceedings of the 42nd International Conference on Machine
  Learning}, volume 267 of \emph{Proceedings of Machine Learning Research}.
  Vancouver, BC, Canada: PMLR.

\bibitem[{Zhong et~al.(2025)Zhong, Li, Xu, Wen, Li, and
  Xu}]{zhong2025solvedetectverify}
Zhong, J.; Li, Z.; Xu, Z.; Wen, X.; Li, K.; and Xu, Q. 2025.
\newblock Solve-Detect-Verify: Inference-Time Scaling with Flexible Generative
  Verifier.
\newblock arXiv:2505.11966.

\bibitem[{Zhou et~al.(2025)Zhou, Xu, Wang, Xiong, and
  Joty}]{zhou2025evaluating}
Zhou, Y.; Xu, A.; Wang, P.; Xiong, C.; and Joty, S. 2025.
\newblock Evaluating Judges as Evaluators: The {JETTS} Benchmark of
  {LLM}-as-Judges as Test-Time Scaling Evaluators.
\newblock In \emph{Proceedings of the 42nd International Conference on Machine
  Learning}, volume 267 of \emph{Proceedings of Machine Learning Research}.
  Vancouver, BC, Canada: PMLR.

\end{thebibliography}

\end{document}